\documentclass{article} \usepackage[final, main, nonatbib]{neurips_2026} \usepackage{hyperref} \usepackage{url} \usepackage{graphicx} \usepackage{booktabs} \usepackage{amsmath,amssymb} \usepackage{microtype} \usepackage[table]{xcolor} \usepackage{multirow} \usepackage{array}
\usepackage{enumitem}

\newcommand{\YesMk}{\ensuremath{\checkmark}}
\newcommand{\NoMk}{--}
\newcommand{\PtMk}{\ensuremath{\circ}}

\graphicspath{{figures_paper/}}

\title{Auditing AI peer reviewers: \\
a benchmark on real scientific papers}

\author{%
  \'I\~nigo Zubeldia$^{1,2}$ \qquad Boris Bolliet$^{3,2}$ \qquad Francisco Villaescusa-Navarro$^{4}$ \\[3pt]
  \textbf{Pablo Villanueva-Domingo$^{5}$} \qquad \textbf{Miles Cranmer$^{1,6}$} \\[5pt]
  \small
  $^{1}$DAMTP, University of Cambridge \quad
  $^{2}$Kavli Institute for Cosmology, University of Cambridge \\
  $^{3}$Cavendish Astrophysics, University of Cambridge \\
  $^{4}$Center for Computational Astrophysics, Flatiron Institute \\
  $^{5}$Computer Vision Center, Universitat Aut\`onoma de Barcelona \quad
  $^{6}$Polymathic AI \\
  \texttt{iz221@cam.ac.uk}
}

\newcommand{\skepthical}{\textsc{Skepthical}}

\begin{document}

\maketitle

\begin{abstract}
LLM-based peer review is now used at production scale, but its evaluation remains limited. We introduce a benchmark that injects objective, localisable errors of four types (general issues, mathematics, numerics, citations) at three severity levels into real papers and measures each reviewer's detection rate by error type and severity, together with its false-positive rate on clean copies. We also present \skepthical{}, a multi-agent reviewer that verifies citation-backed claims by retrieving the cited papers, mathematics with computer algebra, and numerical claims with executed code. We apply the benchmark to two corpora of 15 recent arXiv preprints each, one in astrophysics and one in machine learning (ML), injecting a total of 357 errors, and evaluate twelve reviewer systems: six single-LLM reviewers, two Claude Code configurations, three publicly available reviewers, and \skepthical{}. The choice of model has the largest impact on detection rate: given the same prompt, three frontier LLMs differ by about $50$ percentage points in total detection on both corpora. Tools and scaffolding help when they contribute what the model cannot supply on its own. In particular, on astrophysics, Claude Code improves on a single Claude call in mathematics and numerics detection, and enforced citation retrieval gives \skepthical{} the highest citation error detection rate of all systems by a large margin. This citation lead, however, is not observed on ML, where the models appear to know the literature better and \skepthical{}'s citation detection rate is statistically indistinguishable from that of the best single LLM. False-positive rates are below $10\,\%$ for most systems, but vary with prompt, model, and scaffolding, with mathematics flags the main source of false positives. The evaluation pipeline, the injected errors, and the adjudication logs are released as a benchmark.
\end{abstract}

\section{Introduction} \label{sec:intro}

Peer review is a cornerstone of the scientific process. With AI tools becoming increasingly used in research and AI scientists starting to conduct research at scale~\cite{boiko2023coscientist,lu2024aiscientist,yamada2025aiscientistv2,schmidgall2025agentlab,swanson2025virtuallab,xu2025cmbagent,villaescusanavarro2025denario,hambardzumyan2026aira}, the need for reliable and scalable peer review is greater than ever. In parallel, LLM-based peer review has reached production scale in 2026, with a growing ecosystem of AI reviewers and review venues~\cite{stanford2025,bianchi2025agents4science,idahl2025openreviewer,tu2026paperaudit,biswas2026aaai,darcy2024marg,chang2025treereview,zeng2025reviewrl}. Despite the publication of the first AI-reviewer benchmarks~\cite{li2025llmreval}, methodology for evaluating these AI reviewers has generally lagged behind. Evaluation is most often framed either as alignment with human-written reviews~\cite{liang2023canllm,darcy2024marg,stanford2025} or as synthetic error injection on Machine Learning (ML) papers only~\cite{tu2026paperaudit,biswas2026aaai,xi2025flaws,dycke2025automatic}, with important issues such as false-positive rates or error detection as a function of the error severity not being typically investigated.

This paper addresses these concerns with two primary contributions. First, we introduce
an evaluation pipeline for reviewer systems that injects errors into a paper's \LaTeX{} source at three severity levels (subtle, clear, blatant) across four error types (general issues, mathematics, numerics, citations), and then measures i) the error detection rate for each severity level and error type and ii) the false-positive rates (on a copy with no injected errors). The verdicts are adjudicated by a cross-family LLM panel (Anthropic + OpenAI + Google) following recent LLM-as-judge prescriptions~\cite{panickssery2024judge,verga2024poll,ye2024bias}, with additional mechanical verifiers for mathematics and numerics. Our `dose-response' design extends perturbation-based behavioural evaluation~\cite{ribeiro2020checklist,zhou2024mathcheck} with graded severity levels inspired by~\cite{luo2026meddial}. We note that our benchmark focuses on whether reviewers can detect objective, localisable issues (e.g., numerical and mathematical errors, misrepresented citations, internal inconsistencies), leaving aside questions such as significance and novelty assessment or overall review quality, which are harder to quantify.

Second, we introduce \skepthical{}, an LLM-powered paper reviewer comprising four specialist modules: a citations module that retrieves and reads the cited papers, a mathematics module that verifies mathematical claims using computer algebra, a numerics module that verifies numerical results through executed code, and a general-issues module consisting of an ensemble of cross-family LLMs. Unlike prior reviewers that rely entirely on LLMs, \skepthical{} verifies LLM-flagged issues where possible: the mathematics and numerics modules use computer algebra and executed code, respectively, while the citations module checks the claims directly against the retrieved PDFs of cited papers.

We apply our evaluation pipeline to two corpora of 15 recent astrophysics arXiv preprints and 15 ML preprints, benchmarking \skepthical{} alongside eleven additional reviewer systems: six single-LLM reviewers, two Claude Code configurations, and three publicly available external AI reviewers (the Stanford Agentic Reviewer~\cite{stanford2025}, OpenReviewer-8B~\cite{idahl2025openreviewer}, and the PaperAudit-Bench detector~\cite{tu2026paperaudit}). To our knowledge, our application to the astrophysics corpus is the first controlled comparison of AI peer-reviewer systems on a real, non-ML scientific paper corpus. We release the evaluation pipeline, together with the paper lists, injection catalogues, and full adjudication logs, as a public benchmark at \href{https://github.com/inigozubeldia/skepthical-benchmark}{github.com/inigozubeldia/skepthical-benchmark}.

Table~\ref{tab:related_work} positions this work against prior AI peer-review systems and benchmarks across nine evaluation axes. No prior work grades severity levels or uses a cross-family adjudication panel; only one~\cite{liang2023canllm} covers papers beyond ML/NLP; only one~\cite{dycke2025automatic} provides false-positive control; and only one~\cite{biswas2026aaai} integrates mechanical verifiers. This work is the first to combine all five. Unlike in some previous work, our review adjudication is LLM-based (with additional mechanical checks for mathematics and numerics), although we check the performance of our pipeline against a human annotator on one paper.

\begin{table}[t]
\centering
\footnotesize
\setlength{\tabcolsep}{4pt}
\resizebox{\columnwidth}{!}{%
\begin{tabular}{l c c c c c c c c c}
\toprule
System & Sev & XFamJud & NonML & FP & Mech & ErrType & Scaff & HumEval & Release \\
\midrule
ReviewerGPT~\cite{liu2023reviewergpt}        & \NoMk & \NoMk & \NoMk & \NoMk & \NoMk & \YesMk & \NoMk & \PtMk & \YesMk \\
Liang et al.~\cite{liang2023canllm}          & \NoMk & \NoMk & \YesMk & \NoMk & \NoMk & \YesMk & \NoMk & \YesMk & \PtMk \\
MARG~\cite{darcy2024marg}                    & \NoMk & \NoMk & \NoMk & \NoMk & \NoMk & \YesMk & \YesMk & \YesMk & \YesMk \\
AgentReview~\cite{jin2024agentreview}        & \NoMk & \NoMk & \NoMk & \NoMk & \NoMk & \YesMk & \YesMk & \NoMk & \YesMk \\
TreeReview~\cite{chang2025treereview}        & \NoMk & \NoMk & \PtMk & \NoMk & \NoMk & \NoMk & \YesMk & \YesMk & \YesMk \\
CycleResearcher~\cite{weng2024cycleresearcher} & \NoMk & \NoMk & \NoMk & \NoMk & \NoMk & \NoMk & \PtMk & \YesMk & \YesMk \\
DeepReviewer~\cite{zhu2024deepreviewer}      & \NoMk & \NoMk & \NoMk & \NoMk & \PtMk & \NoMk & \YesMk & \NoMk & \YesMk \\
PaperAudit-Bench~\cite{tu2026paperaudit}     & \NoMk & \NoMk & \NoMk & \PtMk & \PtMk & \YesMk & \YesMk & \NoMk & \YesMk \\
AAAI-26 Pilot~\cite{biswas2026aaai}          & \NoMk & \NoMk & \NoMk & \NoMk & \YesMk & \YesMk & \YesMk & \YesMk & \NoMk \\
FLAWS~\cite{xi2025flaws}                     & \NoMk & \NoMk & \NoMk & \NoMk & \NoMk & \NoMk & \NoMk & \YesMk & \YesMk \\
Dycke \& Gurevych~\cite{dycke2025automatic}  & \NoMk & \NoMk & \NoMk & \YesMk & \NoMk & \NoMk & \YesMk & \YesMk & \YesMk \\
\midrule
\textbf{This work}                           & \YesMk & \YesMk & \YesMk & \YesMk & \YesMk & \YesMk & \YesMk & \PtMk & \YesMk \\
\bottomrule
\end{tabular}%
}
\caption{\textbf{Coverage of nine evaluation axes by prior AI peer-review systems and benchmarks} (\YesMk{} covered; \PtMk{} partial; \NoMk{} not covered. Axes: \textit{Sev} (error severity investigated); \textit{XFamJud} (cross-family judge panel); \textit{NonML} (beyond ML/NLP); \textit{FP} (false-positive evaluation on clean inputs); \textit{Mech} (mechanical verifiers: computer algebra, units, citation lookup); \textit{ErrType} (error-type analysis); \textit{Scaff} (reviewer-scaffolding ablation); \textit{HumEval} (human evaluation); \textit{Release} (public code/benchmark release).}
\label{tab:related_work}
\end{table}

\section{Benchmark design} \label{sec:eval}

\subsection{Reviewer systems} \label{sec:systems} We evaluate the following twelve reviewer systems.

  \begin{itemize}[leftmargin=*,itemsep=2pt,topsep=2pt,parsep=0pt]                                                
  \item \textbf{\skepthical{}} (full pipeline description in Appendix~\ref{app:skepthical}). \skepthical{} is a multi-agent reviewer with four parallel specialist modules: (i) \emph{general issues}, which uses two cross-family LLMs (Claude Opus 4.7 and Gemini 2.5 Pro) to generate two independent reports that are then merged by a third-family model (GPT-5.1), alongside a free-form reading of the paper by a reasoning model (GPT-5.4); (ii) \emph{mathematics}, in which the paper's key mathematical statements are extracted and internally-verifiable claims are verified mechanically with either \texttt{wolframscript} or \texttt{pint} (the former for symbolic equality, the latter for dimensional analysis); (iii) \emph{numerics}, in which the consistency of the numerical statements in the paper is verified with Python code; and (iv) \emph{citations}, in which every key factual statement relying on external references is verified with an LLM with the cited papers provided in its context. To our knowledge, \skepthical{} is the first AI peer reviewer that combines mechanical verification of mathematics and numerics with retrieval-based verification of factual citations. It is publicly accessible at \href{https://skepthical-ai.org}{skepthical-ai.org}. The configuration evaluated below is the public release frozen on 2026-04-21, one day before the 2026-04-22 arXiv submissions from which the astrophysics corpus is sampled (\S\ref{sec:corpus}). No prompt or model choice was tuned in response to the papers or results reported in this paper.

  \item \textbf{Single-LLM reviewers.} We consider two configurations of a single-LLM reviewer: a \emph{free-form} configuration, run with a one-sentence prompt (`\textit{Write a referee report for this scientific paper as if you were a journal referee}'), and a \emph{structured} configuration, run with the full operating instructions of all of \skepthical{}'s agents compiled into one 978-line prompt  (report-merging and tool-output adjudication agents removed and with analogous mechanical-verification-free instructions given for mathematics and numerics). Both configurations are evaluated for three different models (Claude Opus 4.7, GPT-5.5, and Gemini 3.1 Pro Preview, hereafter Gemini 3.1 Pro), giving six single-LLM reviewers in total. We further evaluate a concise version of the structured configuration in Appendix~\ref{app:concise}.

  \item \textbf{Claude Code.} We use the frontier general-purpose coding agent Claude Code \cite{anthropic2025claudecode}, running with Claude Opus 4.7 in a sandbox with the following allowed tools: \verb|Read|, \verb|Bash|, \verb|Grep|, \verb|Glob|, \texttt{python3} (with NumPy, SymPy, SciPy, \texttt{pint}), \texttt{wolframscript}, and arXiv access. As for the single-LLM reviewers, we consider two configurations (sharing model, sandbox, and tools): a \emph{free-form} configuration, run with the same one-sentence referee prompt as in the free-form single-LLM reviewers, and a \emph{structured} configuration, run with a prompt containing \skepthical{}'s full operating instructions.
  
  \item \textbf{Three publicly available external reviewers.} (i) The Stanford Agentic Reviewer at                   
  \href{https://paperreview.ai}{paperreview.ai}~\cite{stanford2025} (closed-source, internals undocumented, queried on 2026-05-01; only run on the astrophysics corpus); (ii)
  \emph{OpenReviewer-8B}~\cite{idahl2025openreviewer}, an open-weights Llama-3.1-8B fine-tuned on $\approx       
  79\,$K~ICLR / NeurIPS reviews, self-hosted and invoked with the published ICLR-template prompt; and (iii) the \emph{PaperAudit-Bench detector}~\cite{tu2026paperaudit}, an open-weights Qwen3-14B-based reviewer, run in
   Deep mode. We do not retrain or re-prompt any of these
  reviewers.
  \end{itemize}

Together, the twelve systems allow for three comparisons and one external reference point. The first comparison is the effect of the \emph{prompt}, as within the single-LLM group and within the Claude Code group, the free-form and structured configurations differ only in the prompt. The second is the effect of the \emph{agentic loop with tool-use}: at each prompt level, the Claude Code configuration can be compared with the Claude Opus~4.7 single-LLM configuration with the same prompt, so the difference between them is not the instructions but the ability to perform several LLM calls, run code, read files, and fetch cited papers. The third is the effect of agentic \emph{scaffolding}: structured Claude Code and \skepthical{} share the same operating instructions, but in \skepthical{} those instructions are executed by dedicated specialist agents that are enforced to run, whereas Claude Code has a much less strict scaffolding. Finally, the three external reviewers show how publicly available systems that an author could use today perform on the same benchmark, and the two open-weights reviewers also test whether small models fine-tuned on review data can compete with frontier systems. Two caveats apply: i) \skepthical{} uses models from all three providers, so its margin over any single-provider system mixes the effect of the pipeline with that of cross-provider ensembling, and ii) none of these comparisons is a capacity-matched ablation of \skepthical{}.  Compute resources and asset licences are summarised in Appendix~\ref{app:compute}.

\subsection{Corpora} \label{sec:corpus}

We evaluate on two corpora of 15 arXiv preprints each, one in astrophysics and one in ML. Each corpus consists of submissions to a single arXiv category on a single recent day, capped at 20 pages to limit cost. The astrophysics corpus comprises 15 of the 24 \texttt{astro-ph} submissions of 2026-04-22 within the page cap (median 13 pages), whereas the ML corpus comprises 15 of the 74 eligible \texttt{cs.LG} submissions of 2026-07-15, drawn by stratified random sampling over eight subfields (median 9 pages, range 3--19). Both chosen days postdate the knowledge cut-off of every model we use. Both paper lists are given in Appendix~\ref{app:corpus}.

Astrophysics was chosen for its abundant formal mathematics and numerics, two key aspects that our benchmark evaluates, and because most existing AI-reviewer evaluations have run on ML papers~\cite{tu2026paperaudit, biswas2026aaai, liang2023canllm, idahl2025openreviewer, darcy2024marg}. ML was added as the second corpus i) to test whether our conclusions hold beyond a single field and ii) because it is the field on which the two fine-tuned open-weights external reviewers were trained, so their performance can be measured in-distribution as well as out of it. We note that no component or prompt in any of the reviewer systems is field-specific.

\subsection{Error injection} 
\label{sec:injection} 

For each paper, four versions are reviewed: one clean copy (used to assess false positives) and three error-injected copies, each carrying one error of each of four types (general issues, mathematics, numerics, citations) at a chosen severity ($S_1$ subtle, $S_2$ clear, or $S_3$ blatant). Across the three injected copies of a paper, every (error type, severity) combination appears exactly once (see Appendix~\ref{app:design}). Spreading the 12 injections across three copies (rather than concentrating them in one) keeps the per-copy error density small, which we take to be more representative of a typical real paper. Each corpus thus leads to 15 clean and 45 error-injected copies, with the astrophysics corpus containing a total of $15 \times 12 - 3 = 177$ injected errors (one paper had too few equations to admit mathematics injections) and its ML counterpart, $180$. 

Representative examples of the three severity levels, by error type, are as follows (full taxonomy with examples in Appendix~\ref{app:taxonomy}). \emph{Citations} run from a 1$\sigma$ value shift misquotation ($S_1$) through a significant methodology misattribution ($S_2$) to a reversed cited conclusion ($S_3$); \emph{numerics} from an abstract--body second-decimal drift ($S_1$) through a digit swap or percent-of error ($S_2$) to a structural inconsistency such as percentages summing above 100\,\% ($S_3$); \emph{mathematics} from a sign flip on a subleading term ($S_1$) through an operator swap or normalisation drop ($S_2$) to a dimensional absurdity ($S_3$); and \emph{general issues} from a typo ($S_1$) through a methods--results inconsistency or phantom-evidence insertion ($S_2$) to an abstract--body contradiction ($S_3$). We note that the taxonomy may not fully reflect the distribution of naturally occurring errors in papers and that severity levels are designed to be calibrated within each error type but not across them, so a given severity label may not represent the same detection difficulty across error types.

Each injection is performed by an agent (Claude Sonnet 4.6 for the astrophysics corpus, Gemini 3.1 Pro for the ML corpus) that takes as input the full paper \LaTeX{} source and returns a suggested edit. The suggested edit is accepted only if it passes a mechanical validator (Appendix~\ref{app:taxonomy}); otherwise an injection of the same type is re-attempted. 

\subsection{Adjudication pipeline} \label{sec:reverif}

Once the reviewer systems have processed all 60 paper copies for each corpus (45 error-injected and 15 clean), we apply an adjudication pipeline to every
resulting review to assess (i) which injected errors were detected on the error-injected copies (\emph{detection matching}), and (ii) which of the issues raised on
the clean copies are genuine false positives (\emph{false-positive           
adjudication}).
  
\textbf{Detection matching.} For each (paper, injected copy, reviewer), the review plus the error-injection records are given to a three-judge cross-family panel (Claude Sonnet 4.6, GPT-5.4, and Gemini 3 Flash Preview, hereafter Gemini 3 Flash). Each judge returns a verdict for each injection ($\{$detected, confidence, matching\_evidence, reason$\}$); the final verdict is the 3-way majority. Low-confidence or 2-1 ambiguous verdicts trigger a PDF-attached escalation pass, in which a single judge (Claude Sonnet 4.6) adjudicates the injection with the full paper PDF in context.

Across the $4{,}084$ per-error-type adjudications in the two corpora ($2{,}104$ astrophysics, $1{,}980$ ML), 93.5\% reach unanimous first-round agreement (3-0) and 6.5\% are 2-1 splits.
The PDF-attached escalation fires on 34.9\% of injections (38.7\% astrophysics, 30.8\% ML), with the dominant trigger being panel uncertainty rather than disagreement. Of the escalated injections, 92.3\% confirm the first-round majority, and 73.6\% return at high confidence. Aggregating across both stages, 97.3\% of all adjudications end with a verdict that was either accepted at first round or confirmed under PDF re-judgement, only 2.7\% are reversed by the escalation pass, and 91.6\% reach unanimous agreement across every judge that voted (first-round panel plus, where applicable, the escalation pass). Only 7.4\% of adjudications retain a low-confidence verdict after escalation. The adjudication is therefore highly self-consistent (judge-level agreement statistics in Appendix~\ref{app:panel-kappa}).

  \textbf{False-positive adjudication on clean copies.} Each review is processed by an agent (GPT-4.1) tasked with identifying clear positives
   (as opposed to low-confidence or open-to-interpretation points) and classifying them into four types: citations, mathematics, numerics, and
  general issues. These detections are then passed into a type-dependent
  pipeline. Citation detections are assessed by a two-judge cross-family panel with the cited PDFs attached (GPT-5 and Claude Sonnet 4.6; if needed, tie-break with Gemini 3.1 Pro). Numerics detections are checked mechanically: GPT-4.1 turns each one into a concrete numerical check to be carried out, a second independent GPT-4.1 call 
  confirms that the numbers involved appear in the paper, and the check is then run in Python. Mathematics detections are checked with \skepthical{}'s mathematics module (which uses \texttt{wolframscript} and \texttt{pint}). Finally, general
  issues detections are adjudicated with a 3-judge cross-family panel with PDF attached (Claude Sonnet 4.6, GPT-5.4 with medium reasoning, and Gemini 3.1 Pro). Outcome verdicts are \textsc{confirmed-issue}, \textsc{fp}, or \textsc{ambiguous}.

\section{Results and Discussion} \label{sec:results}

\subsection{Detection rates of injected errors} \label{sec:res-detection}

\begin{table}[t]
\caption{Error detection rate (as \%) by error type for all the reviewers on the astrophysics corpus (177 injections, top block) and the ML corpus (180 injections, bottom block). Quoted uncertainties are paper-level bootstrap 95\,\% confidence intervals (CIs); bold corresponds to column-best within each block.} \label{tab:headline} \centering \footnotesize \setlength{\tabcolsep}{4pt}
\begin{tabular}{l l c c c c c}
\toprule
 & Reviewer & Citations & Numerics & Mathematics & General & Total \\
\midrule
\multicolumn{7}{l}{\textbf{Astrophysics corpus}} \\
\midrule
\multirow{3}{*}{\emph{\shortstack[l]{Single LLM\\free-form}}}
 & Claude Opus 4.7         & $9.3\pm 10$  & $55.8\pm 15$ & $32.5\pm 17$ & $48.8\pm 15$ & $36.7\pm 8.5$ \\
 & GPT-5.5                 & $22.2\pm 13$ & $80.0\pm 10$ & $73.8\pm 13$ & $66.7\pm 12$ & $60.5\pm 6.0$ \\
 & Gemini 3.1 Pro          & $4.4\pm 5.6$ & $13.3\pm 13$ & $2.4\pm 3.6$ & $26.7\pm 12$ & $11.9\pm 5.8$ \\
\midrule
\multirow{3}{*}{\emph{\shortstack[l]{Single LLM\\structured}}}
 & Claude Opus 4.7         & $20.9\pm 12$ & $79.1\pm 12$ & $55.0\pm 16$ & $58.1\pm 18$ & $53.3\pm 9.1$ \\
 & GPT-5.5                 & $18.2\pm 8.7$& $79.5\pm 12$ & $\mathbf{75.6\pm 12}$ & $72.7\pm 11$ & $61.3\pm 6.6$ \\
 & Gemini 3.1 Pro          & $11.1\pm 7.8$& $57.8\pm 18$ & $45.2\pm 19$ & $35.6\pm 13$ & $37.3\pm 5.7$ \\
\midrule
\multirow{3}{*}{\emph{Agentic}}
 & \skepthical{} & $\mathbf{46.7\pm 13}$ & $80.0\pm 11$ & $73.8\pm 9.5$ & $\mathbf{82.2\pm 8.9}$ & $\mathbf{70.6\pm 6.2}$ \\
 & Claude Code, structured       & $26.7\pm 11$ & $\mathbf{91.1\pm 7.8}$ & $71.4\pm 14$ & $55.6\pm 16$ & $61.0\pm 5.0$ \\
 & Claude Code, free-form        & $8.9\pm 7.8$ & $42.2\pm 18$ & $19.0\pm 11$ & $44.4\pm 12$ & $28.8\pm 6.3$ \\
\midrule
\multirow{3}{*}{\emph{\shortstack[l]{External\\baselines}}}
 & Stanford Agentic Reviewer     & $8.9\pm 10$  & $53.3\pm 12$ & $50.0\pm 19$ & $51.1\pm 11$ & $40.7\pm 6.4$ \\
 & OpenReviewer-8B               & $0.0$        & $0.0$        & $0.0$        & $4.4\pm 5.6$ & $1.1\pm 1.4$ \\
 & PaperAudit-Bench Deep         & $0.0$        & $6.7\pm 8.9$ & $2.4\pm 3.6$ & $0.0$        & $2.3\pm 2.5$ \\
\midrule
\multicolumn{7}{l}{\textbf{Machine-learning corpus}} \\
\midrule
\multirow{3}{*}{\emph{\shortstack[l]{Single LLM\\free-form}}}
 & Claude Opus 4.7 & $13.3\pm 8.9$ & $55.6\pm 12$ & $40.0\pm 13$ & $66.7\pm 14$ & $43.9\pm 6.1$ \\
 & GPT-5.5 & $28.9\pm 14$ & $80.0\pm 10$ & $77.8\pm 14$ & $68.9\pm 13$ & $63.9\pm 7.2$ \\
 & Gemini 3.1 Pro & $0.0$ & $4.4\pm 5.6$ & $11.1\pm 10$ & $28.9\pm 12$ & $11.1\pm 5.8$ \\
\midrule
\multirow{3}{*}{\emph{\shortstack[l]{Single LLM\\structured}}}
 & Claude Opus 4.7 & $57.8\pm 13$ & $84.4\pm 11$ & $77.8\pm 13$ & $68.9\pm 11$ & $72.2\pm 6.1$ \\
 & GPT-5.5 & $57.8\pm 14$ & $\mathbf{91.1\pm 7.8}$ & $\mathbf{91.1\pm 10}$ & $\mathbf{84.4\pm 8.9}$ & $\mathbf{81.1\pm 5.3}$ \\
 & Gemini 3.1 Pro & $26.7\pm 11$ & $57.8\pm 13$ & $62.2\pm 13$ & $57.8\pm 6.7$ & $51.1\pm 5.6$ \\
\midrule
\multirow{3}{*}{\emph{Agentic}}
 & \skepthical{} & $53.3\pm 12$ & $\mathbf{91.1\pm 6.7}$ & $77.8\pm 11$ & $\mathbf{84.4\pm 8.9}$ & $76.7\pm 5.6$ \\
 & Claude Code, structured & $\mathbf{60.0\pm 12}$ & $86.7\pm 14$ & $71.1\pm 16$ & $75.6\pm 13$ & $73.3\pm 8.6$ \\
 & Claude Code, free-form & $15.6\pm 10$ & $42.2\pm 19$ & $24.4\pm 16$ & $55.6\pm 11$ & $34.4\pm 8.6$ \\
\midrule
\multirow{2}{*}{\emph{\shortstack[l]{External\\baselines}}}
 & OpenReviewer-8B & $0.0$ & $2.2\pm 3.3$ & $0.0$ & $4.4\pm 5.6$ & $1.7\pm 1.7$ \\
 & PaperAudit-Bench Deep & $0.0$ & $4.4\pm 5.6$ & $2.2\pm 3.3$ & $4.4\pm 5.6$ & $2.8\pm 1.9$ \\
\bottomrule
\end{tabular}
\end{table}

\paragraph{Detection by error type.} Table~\ref{tab:headline} gives total and per-error-type detection rates on both corpora, with 95\,\% CIs as $\pm$ half-widths from $n_\mathrm{boot} = 2000$ paper-level bootstrap replicates (the scheme used for every interval in this section). In the following, we call a difference between two systems significant when a paired paper-level bootstrap puts less than $5\,\%$ probability on its sign reversing (Appendix~\ref{app:stats}). The two highest totals on each corpus belong to \skepthical{} and structured GPT-5.5: $70.6 \pm 6.2$ and $61.3 \pm 6.6\,\%$, respectively, on astrophysics, a significant difference; and $76.7 \pm 5.6$ and $81.1 \pm 5.3\,\%$ on ML, where the difference is not significant. On \emph{general issues}, the highest rate is \skepthical{}'s on astrophysics ($82.2\,\%$; next is structured GPT-5.5 at $72.7\,\%$, a difference short of significance) and is shared by \skepthical{} and structured GPT-5.5 on ML ($84.4\,\%$ each). On \emph{mathematics}, the highest is structured GPT-5.5's on both corpora, by a small, non-significant margin on astrophysics ($75.6$ vs $73.8\,\%$ for \skepthical{} and free-form GPT-5.5) and a larger one on ML ($91.1$ vs $77.8\,\%$ for \skepthical{}) that falls just short of significance. On \emph{numerics} the highest is structured Claude Code's on astrophysics ($91.1\,\%$, obtained by writing and running ad-hoc Python checks; see below) and is shared by \skepthical{} and structured GPT-5.5 on ML ($91.1\,\%$ each). On \emph{citations}, \skepthical{}, the only system in which PDF retrieval is enforced, is at $46.7\,\%$ against at most $26.7\,\%$ for any other system on astrophysics, whereas on ML structured Claude Opus 4.7, structured GPT-5.5, and structured Claude Code are at $57.8$--$60.0\,\%$ and \skepthical{} at $53.3\,\%$; we examine the reason for this behaviour below. 

\paragraph{External reviewers.} The two ML-tuned open-weights reviewers detect almost no injections on either corpus ($\le 2.3\,\%$ on astrophysics; $1.7$ and $2.8\,\%$ on ML, their home field). The Stanford Agentic Reviewer, run on astrophysics only, reaches $40.7\,\%$, $21$ percentage points (pp) below the strongest single LLM and $30\,\mathrm{pp}$ below \skepthical{}.

\begin{figure}[t]
\centering \includegraphics[width=0.93\linewidth]{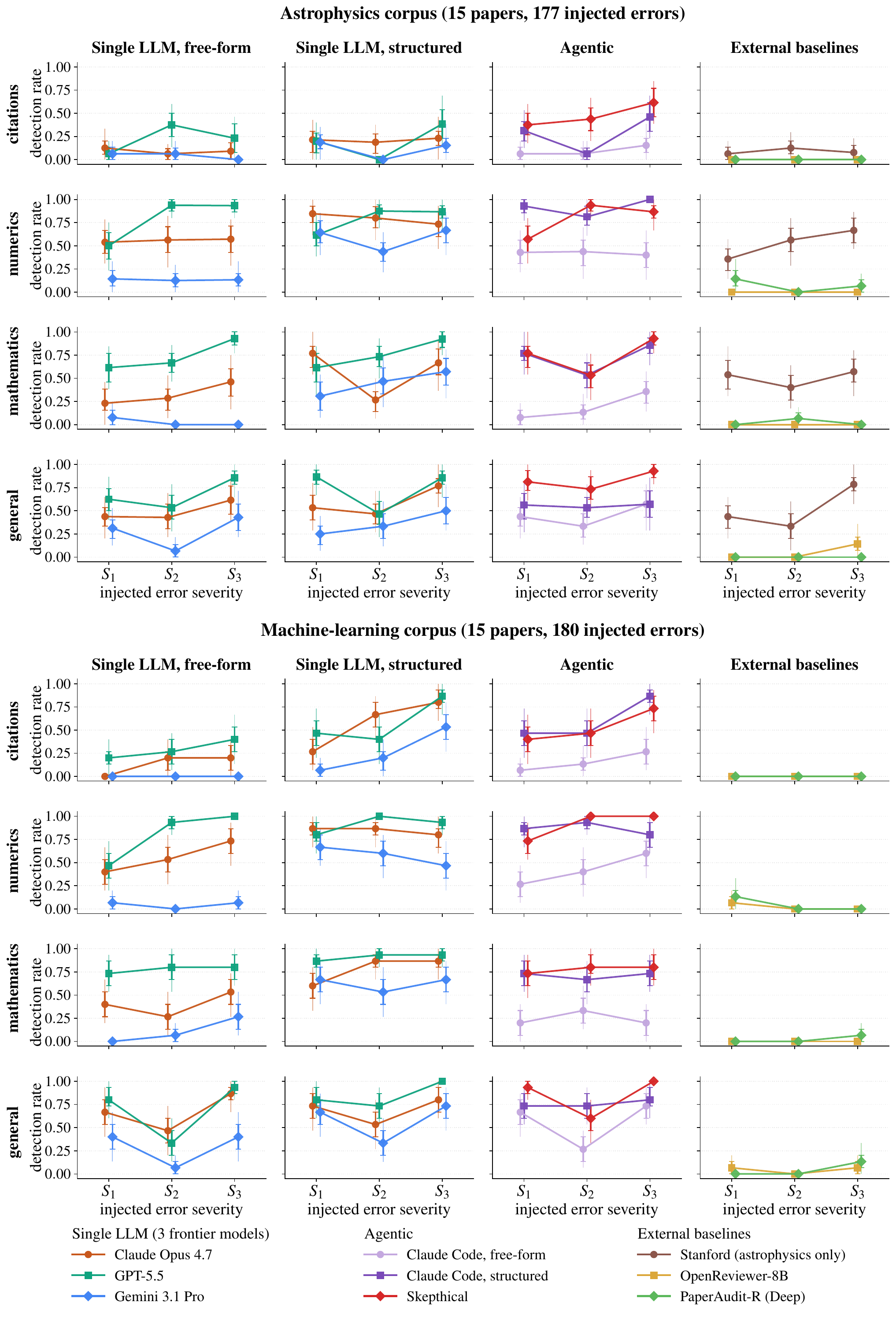} \caption{Error detection rate across error severity, error type, and architectural family for all our reviewers, on the astrophysics corpus (top block) and the ML corpus (bottom block). In each panel, the $x$-axis is the injected error severity, the $y$-axis the detection rate, and the error bars are paper-level bootstrap CIs, with the solid whisker showing the 68\,\% interval and the thinner half-transparent overlay the 95\,\% interval.} \label{fig:doseresponse}
\end{figure}

\paragraph{Detection by severity.} Figure~\ref{fig:doseresponse} adds the third axis, injected severity, on both corpora, with one column per architectural family and one line per provider (single-LLM columns) or per system (agentic and external columns). Detection rises from subtle to blatant in most individual error-type curves and in the average over error types of almost every system. The strongest systems detect a large fraction of the subtle injections ($57$--$73\,\%$ at $S_1$ for \skepthical{}, structured GPT-5.5, and structured Claude Code), whereas the weakest configurations miss most errors even when blatant (the open-weights reviewers below $5\,\%$). The ordering of systems is essentially the same at every severity level, so the comparisons that follow hold across the scale.

\paragraph{Prompt content.} As shown by the per-provider rows of Table~\ref{tab:headline} and per-provider curves of Figure~\ref{fig:doseresponse}, on the same free-form prompt, three frontier LLMs differ by up to $48.6\,\mathrm{pp}$ in total detection on astrophysics (GPT-5.5 $60.5\,\%$, Claude Opus 4.7 $36.7\,\%$, Gemini 3.1 Pro $11.9\,\%$) and by $52.8\,\mathrm{pp}$ on ML ($63.9$, $43.9$, and $11.1\,\%$). Mathematics and numerics drive this discrepancy.

The structured prompt narrows the cross-provider spread considerably by boosting the performance of the weakest models: it has a very large impact for Gemini on both corpora ($+25.4\,\mathrm{pp}$ on astrophysics, $+40.0\,\mathrm{pp}$ on ML) and a smaller but significant one on Claude ($+16.6$ and $+28.3\,\mathrm{pp}$). For GPT-5.5, the effect of the structured prompt is near-negligible on astrophysics ($+0.8\,\mathrm{pp}$), but substantial on ML ($+17.2\,\mathrm{pp}$). In general, we can conclude that detection depends strongly on both the model family and the prompt content

The same prompt change inside the Claude Code group is the cleanest probe of what an operating-instructions prompt causes. With the model (Claude Opus 4.7), sandbox, allowed tools, and arXiv access held fixed, the only difference between the two configurations is the prompt, which moves detection by $+32.2\,\mathrm{pp}$ on astrophysics and $+38.9\,\mathrm{pp}$ on ML. The full output of the astrophysics runs, which we have analysed, shows how the agent uses its available resources. \emph{(i)} The structured agent used tools extensively (median 9 agent turns, max 26), whereas the free-form agent simply read the PDF and replied. \emph{(ii)} The structured agent wrote and ran 21 multi-line Python scripts (using NumPy, SymPy, SciPy, \texttt{math}, and \texttt{mpmath}), whereas the free-form agent wrote none, although it knew it could. \emph{(iii)} Neither used \texttt{wolframscript}, although the structured prompt encourages its use. In summary, with everything else fixed, the prompt alone drives qualitatively different tool use.

\paragraph{Agentic scaffolding.} Holding the structured prompt fixed, going from a single Claude Opus 4.7 call to Claude Code to \skepthical{} takes detection from $53.3$ to $61.0$ to $70.6\,\%$ on astrophysics and from $72.2$ to $73.3$ to $76.7\,\%$ on ML. The first step adds tools at fixed model and prompt, and its gain on astrophysics comes from the error types that a script can check: numerics and mathematics, with $+12.0$ and $+16.4\,\mathrm{pp}$ (both significant); on ML, where there are fewer mathematical and numerical statements to check, there are statistically insignificant changes. 

Going from Claude Code to  \skepthical{} changes more than scaffolding, as \skepthical{} draws on three LLM providers. With this caveat in mind, \skepthical{}'s gains over Claude Code on astrophysics are on citations ($+20.0\,\mathrm{pp}$) and general issues ($+26.7\,\mathrm{pp}$), both significant, while Claude Code is ahead on numerics (by $11.1\,\mathrm{pp}$, significant); on ML, no difference is significant. The citation result is the clearest case for enforcing a verification step rather than offering one: Claude Code had the same arXiv access, but the execution traces show paper retrieval never took place, whereas in \skepthical{} retrieval is mandatory. On astrophysics this puts \skepthical{} $20\,\mathrm{pp}$ ahead of structured Claude Code on citations; on ML it does not ($53.3$ vs $60.0\,\%$, not significant). A possible explanation, developed in the next paragraph, is that the models are familiar enough with the ML literature to judge many cited claims without retrieval from the source, so that enforced retrieval adds little there. 

Finally, compared with the best single-LLM reviewer, structured GPT-5.5 (the highest total among the six single-LLM configurations on both corpora), \skepthical{} is $+9.3\,\mathrm{pp}$ on astrophysics (significant) and $-4.4\,\mathrm{pp}$ on ML (not significant). The multi-agent pipeline therefore  outperforms the best single-LLM reviewer only on the corpus where its enforced steps make a difference, citation retrieval on astrophysics in particular; on ML the two are statistically indistinguishable on overall detection rate. We note, however, that against the concise structured GPT-5.5 of Appendix~\ref{app:concise}, which reaches $68.9\,\%$ on astrophysics and $76.1\,\%$ on ML, the margin is $+1.7$ and $+0.6\,\mathrm{pp}$, neither significant.

\paragraph{Citation detection across fields.} On ML, the structured single LLMs have a higher citation detection rate than on astrophysics (all three providers; GPT-5.5 from $18$ to $58\,\%$), while \skepthical{} barely changes ($47$ to $53\,\%$). We attribute this to two factors, as discussed in more detail in Appendix~\ref{app:citations-transfer}. First, more of the ML citation errors can be detected by having a general knowledge of the field, e.g., a well-known method attributed to the wrong paper (14 of 45, against 4 of 45 on astrophysics). Second, the models appear to know the ML literature better than the astrophysics one, since they also detect more of the errors that we classified as needing retrieval from the cited paper. \skepthical{}, which reads the cited paper in every case, changes little. In conclusion, retrieval makes a difference when the model does not already know the literature.

\paragraph{Mechanical verification.} The detection results understate the contribution of the mechanical verifiers. A frontier LLM matching \skepthical{}'s $80\,\%$ numerics detection rate is not the same finding as \skepthical{} delivering $80\,\%$ recall through tool-mediated deterministic computation. \skepthical{}'s checks invoke deterministic tools (Python calculations, \texttt{wolframscript}, \texttt{pint}, and arXiv-PDF retrieval) on every checked claim and are auditable in source, while the single-LLM equivalent is a single forward pass whose correctness on any given equation is opaque and non-reproducible. Claude Code with the structured prompt sits between the two: it has the same tool affordances as \skepthical{}, but only invokes them on a fraction of instances (and never \texttt{wolframscript}, despite the prompt's encouragement), so its detections are mechanically grounded only on the instances where the agent chose to run a verification script.

\paragraph{Further results.} Six further analyses are reported in the appendices: a test for judge--reviewer family overlap (Appendix~\ref{app:family-pairing}); a human check of the detection adjudication for one paper (Appendix~\ref{app:human}); a concise structured-prompt configuration (Appendix~\ref{app:concise}); the attribution of \skepthical{}'s detections to its modules (Appendix~\ref{app:attribution}); a test of whether errors in the same copy suppress each other's detection (Appendix~\ref{app:severity-coupling}); and the agreement between reviewer systems on which errors they detect (Appendix~\ref{app:inter-reviewer}).

\subsection{False positives} \label{sec:res-precision}

\begin{figure}[t]
\centering \includegraphics[width=0.76\linewidth]{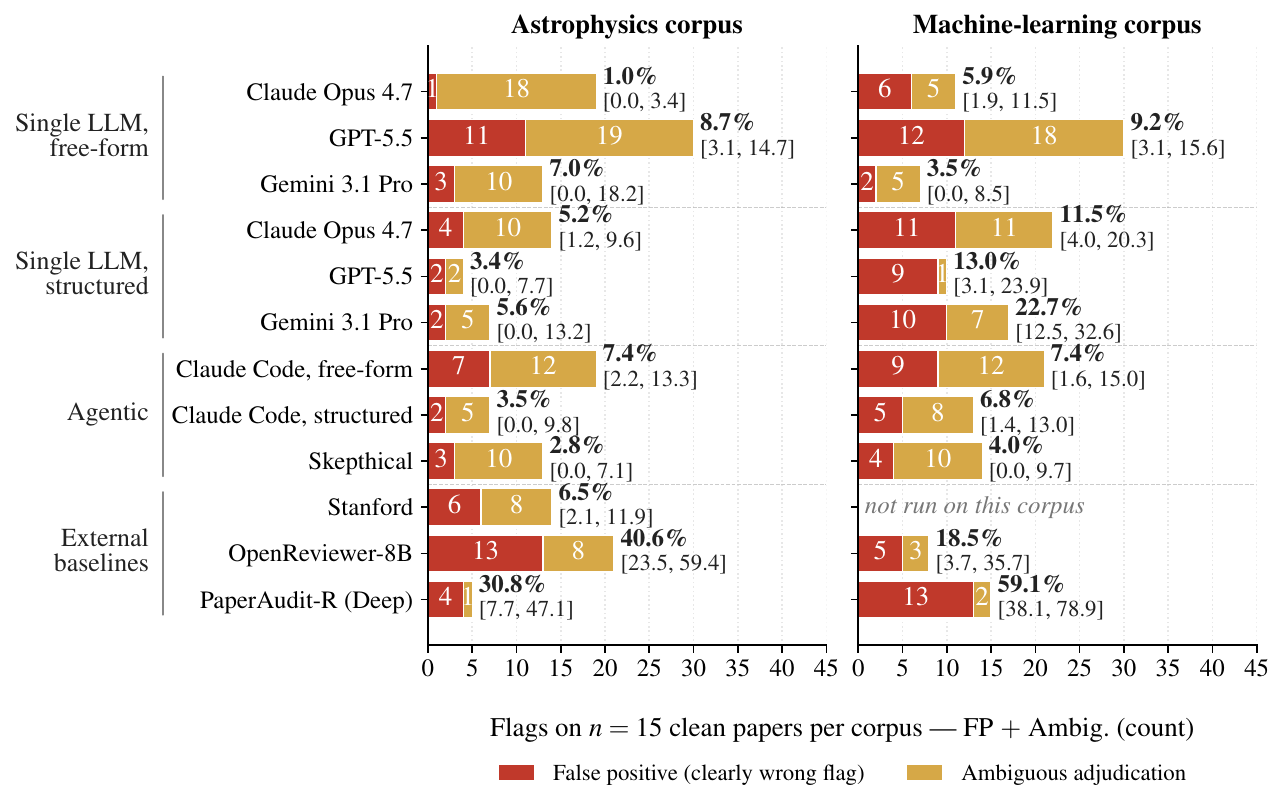} \caption{False-positive and ambiguous flags per system on the 15 clean copies of each corpus (left: astrophysics; right: ML). Each row stacks two bars: clearly-FP flag count (red) and ambiguous-adjudication count (yellow). The percentage at the end of each bar is $\mathrm{FP\,rate} = N_{\textsc{fp}}/N_\mathrm{tot}$, with $N_\mathrm{tot}$ the total number of flags raised on the 15 clean copies of that corpus (definition and panel verdicts in \S\ref{sec:reverif}); the bracketed values are paper-level bootstrap 95\,\% CIs ($n_\mathrm{boot}=2000$). Note that cells with $N_\mathrm{tot} < 5$ on a single error type yield unreliable FP rates.} \label{fig:fp}
\end{figure}

We define a reviewer's false-positive (FP) rate as the fraction of the flags it raises on the 15 clean copies of a corpus that the adjudication panel judges clearly false (\S\ref{sec:reverif}). Figure~\ref{fig:fp} shows, per system and corpus, the FP and \textsc{ambiguous} counts and the FP rate with its paper-level bootstrap 95\,\% CI. Note the rate depends on the count and the total flag count, so a short bar can carry a high rate and vice versa. 

False-positive rates are generally low: apart from the two open-weights reviewers, every system is below $9\,\%$ on astrophysics, and six of nine are below $10\,\%$ on ML, the exceptions being the three structured single LLMs ($11.5$--$22.7\,\%$). \skepthical{} has a low FP rate on both corpora, $2.8\,\%$ on astrophysics (106 flags) and $4.0\,\%$ on ML (99 flags), against $3.4$--$22.7\,\%$ for the structured single LLMs. The lowest rates overall belong to free-form single LLMs, which raise almost no mathematics or numerics flags: Claude Opus 4.7 on astrophysics ($1.0\,\%$) and Gemini 3.1 Pro on ML ($3.5\,\%$).

The effect of the structured prompt on the FP rate is mixed: on astrophysics it narrows the spread across providers, lowering GPT-5.5's rate ($8.7 \to 3.4\,\%$) and raising Claude's ($1.0 \to 5.2\,\%$); on ML it raises every provider's rate; and for Claude Code it lowers the rate on both corpora. Three things, however, are consistent across reviewers and corpora. First, mathematics flags are the least reliable: pooled over the frontier systems, they are judged false at $30\,\%$ on astrophysics and $67\,\%$ on ML, against $3\,\%$ for general-issue flags, and they account for half of all false positives on both corpora. Second, the structured prompt roughly halves the number of general-issue flags, which are rarely wrong, so it can raise the FP rate even when the number of false positives falls (GPT-5.5 on ML: $12 \to 9$ false positives on $131 \to 69$ flags). Finally, the two open-weights detectors perform poorly on both corpora, featuring the highest rates.

\section{Limitations} \label{sec:disc}

The benchmark measures one part of peer review: whether a reviewer detects objective, localisable defects (citation support, mathematical and numerical correctness, internal consistency) and whether it raises false alarms on clean papers. The benchmark does not address issues such as novelty, significance, adequacy of experiments, missing controls, interpretive judgement, or alignment with human-written reviews, as they are significantly harder to quantify and, in general, lack a clear ground truth. Therefore, our results should not be interpreted as holistic measurements of review quality.

Each corpus is small: 15 papers from one day in one field, capped at 20 pages. The rates reported here are therefore estimates for
those two slices, and longer manuscripts are not represented. As already noted, severity labels are calibrated within each error type but not across them. In addition, each injected copy carries four errors, and a blatant one might draw attention away from a subtle one in the same copy. Because severities were assigned independently for each error type, the severity of the other errors in a copy varies at random.  We find that whether an error is detected does not depend measurably on how severe its neighbours are (Appendix~\ref{app:severity-coupling}). However, this test cannot ascertain whether the mere presence of other errors has an effect; that would require copies carrying a single error, which was beyond the cost constraints of this work.

Adjudication is LLM-based for detection matching and for general-issue and citation false positives. Due to resource constraints, the blinded human check covers just one paper and detection only (Appendix~\ref{app:human}). The comparisons involving \skepthical{} are not capacity-matched: it uses more calls and three providers, and we did not run a same-provider ensemble of equal cost, so its gains cannot be attributed to the pipeline design alone. 

For the external reviewers, two specific fairness caveats apply. The Stanford reviewer is hard-capped at the first 15 pages of the PDF; the same page truncation is applied to OpenReviewer-8B's input; and PaperAudit-Bench Deep mode chunks natively and is given the full paper.

Exact reproduction of any system here requires the frozen commercial model versions, which will be deprecated. The benchmark, however, remains reusable, since the evaluation pipeline, the paper lists, the injections, and the adjudication records are released. Finally, three risks accompany deployment of tools like \skepthical{} at scale: false confidence from uneven per-error-type detection, adversarial gaming of known weaknesses, and concentration of review infrastructure on a few LLM providers. The design choices here (cross-family ensembling, auditable verification of mathematics, numerics, and citations) mitigate these risks but do not remove them.

\section{Conclusion} \label{sec:conclusion}

In this paper we have introduced an evaluation pipeline for AI peer reviewers that injects errors of four types and three severity levels into real papers and measures the detection rate as a function of error type, severity, prompt, and reviewer architecture, together with the false-positive rate on clean copies. We have applied it to two corpora of 15 real arXiv preprints each, one in astrophysics and one in ML, and benchmarked twelve reviewer systems: six single-LLM reviewers, two Claude Code configurations, three publicly available external reviewers, and \skepthical{}. \skepthical{} is a multi-agent reviewer, also introduced here, with a reference-attached citation verifier, a computer-algebra-backed mathematics auditor, a code-backed numerical auditor, and a cross-family general-issues ensemble. The evaluation pipeline, together with the injection catalogues, and the full adjudication logs, is released as a benchmark for the AI-reviewer landscape at \href{https://github.com/inigozubeldia/skepthical-benchmark}{github.com/inigozubeldia/skepthical-benchmark}, and \skepthical{} is publicly available at \href{https://skepthical-ai.org}{skepthical-ai.org}.

Our main conclusions are as follows.
\begin{enumerate}[leftmargin=*,itemsep=0pt,topsep=0pt,parsep=0pt,partopsep=0pt]
\item The choices of model and prompt make a difference. Given the same one-sentence prompt, three frontier LLMs differ by about $50\,\mathrm{pp}$ in total detection on both corpora, mostly on mathematics and numerics. The longer structured prompt narrows this spread by lifting the weaker models. For Claude Code, with model and tools fixed, the same prompt change alone triggers extensive tool use and raises detection by over $30\,\mathrm{pp}$ on both corpora.
\item Tools and scaffolding help when they contribute what the model cannot supply on its own. In particular, on astrophysics, Claude Code improves significantly on a single Claude call in mathematics and numerics detection, and enforcing citation retrieval gives \skepthical{} a large citation lead there. On ML, however, where the models appear to know the literature better, enforced retrieval does not lead to a significant improvement.
\item Detection rises with severity for every capable system, whereas the two fine-tuned open-weights reviewers have a negligible detection rate.
\item False-positive rates are generally low, below $10\,\%$ for most systems apart from the open-weights reviewers, but depend on prompt, model, and scaffolding. Mathematics flags are the least reliable and account for half of all false positives; \skepthical{} keeps a low rate on both corpora.
\end{enumerate}

Beyond the AI-reviewer landscape itself, peer review remains a cornerstone of the scientific process for both human-authored and increasingly AI-conducted research~\cite{boiko2023coscientist,lu2024aiscientist,yamada2025aiscientistv2,schmidgall2025agentlab,swanson2025virtuallab,xu2025cmbagent,villaescusanavarro2025denario,hambardzumyan2026aira}. In this regard, we see \skepthical{}, and tools like it, both as standalone reviewing infrastructure for human-authored research and as components that AI scientists could integrate as an internal self-audit layer guiding autonomous research.

\begin{ack}
The authors would like to thank Antony Lewis, Licong Xu, Thomas Borrett, and Raul Jimenez for useful discussions. IZ and BB thank the Infosys--Cambridge AI Centre for support. This work was partially funded by an unrestricted gift from Google to BB. Part of the development of \skepthical{} was carried out while IZ was at the Kavli Institute for Theoretical Physics (KITP); his stay was supported in part by grant NSF PHY-2309135 to the KITP. MC is supported by an AI2050 Early Career Fellowship from Schmidt Sciences and by the Isaac Newton Trust; Polymathic AI is funded by the Simons Foundation and Schmidt Sciences. The authors declare no competing interests.
\end{ack}

\clearpage \appendix 
\section{\skepthical{} pipeline detail} \label{app:skepthical}

\begin{figure}[t]
\centering \includegraphics[width=0.99\linewidth]{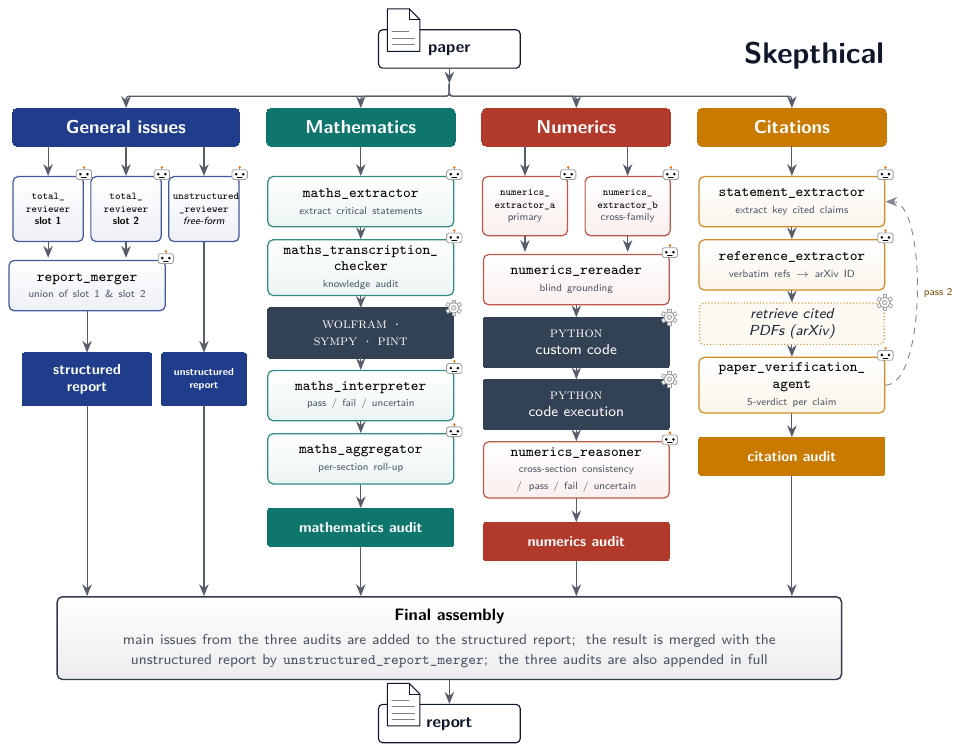}
\caption{The \skepthical{} pipeline. The paper PDF is processed in parallel by four specialist modules (general issues, mathematics, numerics, citations); each module is a short chain of LLM agents (robot icons) interleaved with mechanical or deterministic engines (gear icons). Module outputs converge in a final assembly step that produces the markdown review: main issues from the three audits are added to the structured general-issues report, the result is merged with the unstructured report by \texttt{unstructured\_report\_merger}, and the three audits are also appended in full.}
\label{fig:skepthical}
\end{figure}

\skepthical{} reviews a paper with a pipeline of specialised agents rather than with a single LLM call. The paper is processed in parallel by four modules, one each for general issues, mathematics, numerics, and citations, and their outputs are merged at the end into a single structured review (Figure~\ref{fig:skepthical}). Table~\ref{tab:models} lists the model used by every agent in the configuration run in this work.

\subsection{General-issues module} \label{app:skepthical-general}

The general-issues module produces the backbone of the review. It has two parts: a structured stage, in which two LLMs from different families are combined so that the issues raised do not depend on a single family, and a parallel free-form reading of the paper by a single reasoning model.

The first stage is the \texttt{total\_reviewer} ensemble: two parallel calls, one on Claude Opus 4.7 and one on Gemini 2.5 Pro, each returning a summary, a list of strengths, and three lists of issues by severity (\textsc{major}, \textsc{minor}, and \textsc{very minor}) in a fixed JSON format. Both calls use the same operating instructions, which require a coverage checklist of five categories (general, logical consistency, evidence and citations, presentation and clarity, reproducibility), a consistency pass across sections on named choices and re-used quantities, an audit of rhetoric and scope intended to catch sweeping generalisations and contradictions in direction or kind, and a phantom-evidence audit in which every reference to a figure, table, or other reported result must be matched to the corresponding object in the paper. The two outputs are then merged by \texttt{report\_merger} (GPT-5.1), which removes duplicate issues, combines the strengths, and harmonises the severity levels across the two calls. The merger belongs to a third family (OpenAI), so that neither of the two input families is favoured at the merge step~\cite{panickssery2024judge}.

In parallel with the structured ensemble, the \texttt{unstructured\_reviewer} agent (GPT-5.4, high reasoning) reads the paper without a checklist. Its purpose is not coverage, which the structured pass already provides, but to bring extended reasoning to a workflow whose other parts run without it. This serves two ends. First, the free-form pass can raise concerns about the paper as a whole that fall outside the checklist. Second, it helps to establish the relative importance of the issues found by the structured calls, which are not necessarily reported in order of importance. This pass runs on the third frontier family (OpenAI), so that, taken together, the structured and unstructured parts of the general-issues module span Anthropic, Google, and OpenAI.

\subsection{Mathematics module} \label{app:skepthical-maths}

The mathematics module audits the paper's mathematics with a five-stage pipeline. The first stage extracts the mathematical statements in the paper, the second cross-checks them, the third verifies them mechanically (if applicable), and the fourth and fifth interpret and aggregate the results.

In stage 1, the \texttt{maths\_extractor} agent (Gemini 2.5 Pro, high reasoning) extracts every mathematical statement that the paper presents as central (e.g., an equation, a derivation chain, a limit or series identity, a dimensional or normalisation constraint, a definition, or an inequality) into a structured record. The record contains the \LaTeX{} of the statement, the verbatim source snippet, its location, the kind of statement, a flag marking whether units matter, and an executable plan for checking it. When a derivation chain contains an asserted equation, that equation is also extracted on its own, so that the derivation steps go to mechanical verification while the asserted starting equation receives a separate knowledge audit. The extractor must record every numbered equation it does not extract in an \texttt{excluded\_equations} list, with a reason. If that list is empty while some equations remain uncovered, the extraction is repeated with reinforced instructions.

Stage 2 is a cross-family check of the extracted statements. For each of them, the \texttt{maths\_transcription\_checker} agent (Claude Opus 4.7) verifies two things. First, that the code prepared for the mechanical check in stage 3 faithfully reproduces the statement as it appears in the paper. Second, for any equation that the paper asserts rather than derives, such as a definition or a standard result quoted from the literature, that the equation matches the canonical form of the quantity it describes. This second check targets errors in prefactors, exponents, scaling factors, and subleading terms, which a mechanical check of the paper's own derivation cannot detect, since the derivation can be internally consistent while starting from a wrong equation. We note that stages 1 and 2 run on different model families, which makes it less likely that an error of the extracting model, for instance a misread sign or exponent, is shared by the model that checks its work.

In stage 3, the statements are verified mechanically. Symbolic equalities, limits, series identities, and solvability queries are checked with \texttt{wolframscript}. When a statement contains free symbols, it is also evaluated at ten random points of their parameter space. SymPy serves as a fallback if \texttt{wolframscript} crashes. In addition, \texttt{pint} performs the dimensional analysis of the statements flagged as unit-critical in stage 1. Each check returns one of six outcomes: \textsc{confirmed}, \textsc{refuted}, \textsc{inconclusive}, \textsc{error}, \textsc{skipped}, or \textsc{unavailable}.

In stage 4, the \texttt{maths\_interpreter} agent (Claude Opus 4.7) combines the outcome of the stage-2 check and the stage-3 mechanical verdict for each statement into a final judgement of \textsc{pass}, \textsc{fail}, or \textsc{uncertain}. Finally, in stage 5, the \texttt{maths\_aggregator} agent (Gemini 2.5 Pro, high reasoning) compiles the per-statement judgements into the mathematics audit and writes a mathematics audit report that is attached to the final review.

\subsection{Numerics module} \label{app:skepthical-numerics}

The numerics module verifies numerical statements with code. First, two extractor agents from different model families, \texttt{numerics\_extractor\_a} (Claude Opus 4.7) and \texttt{numerics\_extractor\_b} (GPT-5.5, high reasoning), read the PDF independently and each returns a list of numerical statements to be verified. Every statement is labelled with one of eleven kinds of check (agreement between abstract and body, a table-row total, a table-column total, percentages that should sum to one hundred, a percentage of a stated quantity, a ratio, a difference, an integer equality, a constant repeated across the paper, a rounded value, and a sum that should equal a stated total). To make sure that the statement refers to a number that actually appears in the paper, the extractor must also record where it found it. It does so by quoting a verbatim passage of at least eight consecutive words from the paper that contains the number, together with the number itself exactly as printed, without units or \LaTeX{} markup.

Second, a deterministic Python step compares the two lists and keeps only the statements on which both extractors agree, that is, those for which the quoted passage, the kind of check, and the printed number all match. Statements found by only one extractor are set aside in a separate list of unverified items. Each surviving statement is then re-read by the \texttt{numerics\_rereader} agent (GPT-4.1). This agent is given only the quoted passage and the kind of check, not the number reported by the extractors, and is asked to state which number is printed in that passage. The statement is accepted for verification only if the number the re-reader reports matches the one both extractors gave. This removes a whole class of errors in which an extractor misreads or invents a number.

Third, the accepted statements are verified by Python code, with no LLM involved. For each of the eleven kinds of check there is a fixed function that performs the corresponding arithmetic on the extracted numbers, for instance summing a table row and comparing the result with the printed total. When a check fails, a fixed rule assigns the severity according to the kind of check and the size of the discrepancy. Some problems cannot be found in this way, and a single LLM call is used for them. The \texttt{numerics\_reasoner} agent (GPT-5.5) reads the full PDF once, with the results of the mechanical checks given to it so that it does not repeat them, and looks for two further kinds of problem: a quantity that is given different values in distant parts of the paper, and numerical errors that both extractors missed. Its findings are marked in the audit report as not mechanically verified. Each of them must be anchored to a verbatim quote from the paper, and a finding whose quote cannot be found in the paper is discarded. Finally, the \texttt{numerics\_reporter} agent (GPT-4.1) compiles the mechanical and reasoner findings into a numerics audit report that is attached to the final review.

\subsection{Citations module} \label{app:skepthical-citations}

The citations module checks the factual statements in the paper that rely on external references against the cited papers themselves, which it retrieves. It runs in two passes. In the first, the \texttt{statement\_extractor} agent (GPT-5.1) reads the full text of the paper and returns the $N$ most important such statements. $N$ grows with the length of the paper, at one statement per ten thousand characters, up to a maximum of ten in order to bound the cost. Each statement is returned together with the reference identifier exactly as it appears in the paper (a BibTeX key, a numeric label, or an author-year string). Two further agents then connect the statements to the cited papers. First, the \texttt{reference\_extractor} agent (GPT-4.1) copies the entries of the paper's reference list verbatim. Then, the \texttt{reference\_resolver} agent (GPT-5 mini) maps each in-text identifier to its entry in the reference list and from there to an arXiv identifier, resolving differences in numbering and in author-year forms (a smaller model suffices for this lookup step). The PDFs of the cited papers are then downloaded from the arXiv using the identifiers obtained in this way.

The statements are then verified. For each of them, the \texttt{paper\_verification\_agent} agent (Gemini 2.5 Pro, high reasoning) receives the statement together with the PDF of the cited paper and returns one of five verdicts: \textsc{supported}, \textsc{weakly supported}, \textsc{partially supported}, \textsc{contradicted}, or \textsc{unrelated}. The agent is instructed to use only the attached paper, which limits the influence of what the model may already know, or believe, about the cited work.

In the second pass, the \texttt{statement\_extractor} agent is run again, this time with the first-pass statements and their verdicts given to it so that it does not repeat them, and asked for up to three further statements that look suspicious. These are verified in the same way by the \texttt{paper\_verification\_agent} agent, and their verdicts are labelled as coming from the second pass, so that the two passes can be weighted separately later in the pipeline.

\subsection{Coordination, assembly, and deployment}

The four modules run concurrently, on four worker threads inside the \skepthical{} worker process. Once all four have finished, an assembly step builds the review in a fixed order as follows: it merges the two \texttt{total\_reviewer} outputs through the \texttt{report\_merger} agent, adds the paper title, appends the mathematics and numerics audit reports, folds in the free-form pass through the \texttt{unstructured\_report\_merger} agent, and finally inserts the per-statement verdicts of the citations module. The assembled review is then compiled to markdown and PDF.

\skepthical{} is deployed on Google Cloud Run and is accessible at \href{https://skepthical-ai.org}{skepthical-ai.org}. Submitted papers are placed in a queue and processed by worker instances that run in parallel, so throughput grows with the number of workers. A typical 12-page astrophysics paper takes approximately 24~minutes.

\begin{table}[h]
\caption{Per-agent model assignment for the \skepthical{} multi-agent pipeline as evaluated in this paper. Each row names an agent introduced in the prose above, its role, and the model it runs on. Mechanical engines (Wolfram Engine, SymPy, \texttt{pint}) and pure-Python steps (numerics reconciliation, predicate adjudication, deterministic assembly) are not LLM-driven and are listed separately at the bottom. Models used in the adjudication pipeline (detection matching and false-positive adjudication) are described in \S\ref{sec:reverif} of the main text.} \label{tab:models} \centering \footnotesize \setlength{\tabcolsep}{4pt}
\begin{tabular}{l l l}
\toprule
Agent & Role & Model \\
\midrule
\multicolumn{3}{l}{\emph{General issues}} \\
\texttt{total\_reviewer} (slot 1)  & coverage-checklist referee                & Claude Opus 4.7 \\
\texttt{total\_reviewer} (slot 2)  & coverage-checklist referee (cross-family) & Gemini 2.5 Pro \\
\texttt{report\_merger}            & cross-family merger of slot 1 + slot 2    & GPT-5.1 \\
\texttt{unstructured\_reviewer}    & free-form prose pass                      & GPT-5.4 (high) \\
\texttt{unstructured\_report\_merger} & final assembly merger                  & Gemini 2.5 Pro (high) \\
\midrule
\multicolumn{3}{l}{\emph{Mathematics}} \\
\texttt{maths\_extractor}          & Stage 1: critical-statement extractor     & Gemini 2.5 Pro (high) \\
\texttt{maths\_transcription\_checker} & Stage 2: transcription + knowledge audit & Claude Opus 4.7 \\
\texttt{maths\_interpreter}        & Stage 4: per-statement interpreter        & Claude Opus 4.7 \\
\texttt{maths\_aggregator}         & Stage 5: section-level aggregator         & Gemini 2.5 Pro (high) \\
\midrule
\multicolumn{3}{l}{\emph{Numerics}} \\
\texttt{numerics\_extractor\_a}    & primary extractor                         & Claude Opus 4.7 \\
\texttt{numerics\_extractor\_b}    & cross-family extractor                    & GPT-5.5 (high) \\
\texttt{numerics\_rereader}        & blind location-cue re-reader              & GPT-4.1 \\
\texttt{numerics\_reasoner}        & cross-section consistency reasoner        & GPT-5.5 \\
\texttt{numerics\_reporter}        & prose paraphraser of audit report         & GPT-4.1 \\
\midrule
\multicolumn{3}{l}{\emph{Citations}} \\
\texttt{statement\_extractor}      & key-statement extractor (Pass 1 + Pass 2) & GPT-5.1 \\
\texttt{reference\_extractor}      & verbatim bibliography pull                & GPT-4.1 \\
\texttt{reference\_resolver}       & in-text label $\to$ arXiv-ID lookup       & GPT-5 mini \\
\texttt{paper\_verification\_agent} & PDF-attached 5-verdict claim verifier    & Gemini 2.5 Pro (high) \\
\midrule
\multicolumn{3}{l}{\emph{Mechanical / non-LLM components}} \\
Wolfram Engine via \texttt{wolframscript} & Mathematics symbolic engine & --- \\
SymPy                                  & Mathematics fallback symbolic engine           & --- \\
\texttt{pint}                          &  Dimensional analysis              & --- \\
Python & Numerics verification             & --- \\
\bottomrule
\end{tabular}
\end{table}

\section{Corpus} \label{app:corpus}

Both corpora are listed in Table~\ref{tab:corpus}, which gives each paper's primary category; 6 of the astrophysics papers and 8 of the ML papers are cross-listed to the corpus category. Of the 60 \texttt{astro-ph} submissions of 2026-04-22, 24 were within the 20-page cap, of which 7 could not be processed by the injection toolchain and 2 were injected but not evaluated, leaving the 15 evaluated papers.

Of the 151 \texttt{cs.LG} submissions of 2026-07-15 (cross-lists included), 81 were within the page cap and 74 had \LaTeX{} source that compiled unmodified and contained enough material for all four error types, as judged by an LLM from the PDF (Claude Sonnet). The 15 evaluated papers were drawn from these 74 by stratified random sampling over eight subfields assigned from the titles alone, with quotas of one to three papers per subfield. This screening was not infallible: two drawn papers turned out to have no displayed mathematics and were replaced by the next paper in their subfield. The ML papers are shorter than the astrophysics ones (median 9 against 13 pages) and contain fewer displayed equations (median 5 against 12).

\begin{table}[h]
\caption{The two 15-paper evaluation sets, with arXiv identifier, primary category, page count, and title.} \label{tab:corpus} \centering \footnotesize
\begin{tabular}{l l c p{7.6cm}}
\toprule
arXiv ID & Primary category & Pages & Title \\
\midrule
\multicolumn{4}{l}{\textbf{Astrophysics corpus}} \\
\midrule
\texttt{2604.20080}~\cite{slater2026mrk996} & astro-ph.GA & 14 & Cold molecular gas distribution and kinematics in the low-metallicity dusty starburst of Mrk 996 resolved with ALMA \\
\texttt{2604.20086}~\cite{carrillo2026dsphdm} & hep-ph      & 18 & Photon and neutrino fluxes from spheroidal dwarf galaxies in a decaying DM model \\
\texttt{2604.20142}~\cite{chen2026grb} & astro-ph.HE & 18 & Fast-Cooling Synchrotron in Decaying Magnetic Fields: Implications for the GRB Spectral Distribution \\
\texttt{2604.20203}~\cite{chen2026subneptunes} & astro-ph.EP & 10 & Sub-Neptunes Show a Stronger Correlation with Cold Jupiters than Super-Earths Especially in Metal-rich Systems \\
\texttt{2604.20262}~\cite{geng2026tianlai} & astro-ph.IM & 15 & Measuring Tianlai's primary beam using sky model \\
\texttt{2604.20344}~\cite{baes2026core} & astro-ph.GA & ~8 & The core of the problem: Physical limits of the core-S\'ersic model \\
\texttt{2604.20411}~\cite{he2026sfr} & astro-ph.GA & 14 & The evolution of the SFR--$M_\star$ relation at $0.1 < z < 4$: environmental and morphological dependencies \\
\texttt{2604.20725}~\cite{matsuo2026crust} & nucl-th     & 16 & Interaction between nuclear clusters and superfluid phonons in the neutron-star inner crust \\
\texttt{2604.20809}~\cite{kinney2026geodesic} & gr-qc       & ~8 & Geodesic completeness in general cosmological scenarios \\
\texttt{2604.20914}~\cite{wang2026ringdown} & gr-qc       & 13 & Cracking gravitational-wave multiple ringdown modes in space \\
\texttt{2604.20971}~\cite{garg2026emri} & astro-ph.GA & 12 & Chaotic migration of LISA extreme mass-ratio inspirals in a turbulent accretion disk: effect on waveform de-phasing \\
\texttt{2604.20975}~\cite{quelquejay2026pta} & astro-ph.HE & 10 & Probing supermassive black-hole mergers with pulsar timing arrays \\
\texttt{2604.21007}~\cite{kakimoto2026deepdive} & astro-ph.GA & 19 & DeepDive: simultaneous formation of massive quiescent galaxies in high-redshift galaxy proto-clusters \\
\texttt{2604.21077}~\cite{lella2026axions} & hep-ph      & ~9 & Multimessenger probes of axions from compact objects \\
\texttt{2604.21099}~\cite{padgett2026uhecr} & hep-ph      & 13 & Ultra-high energy cosmic rays from the local void \\
\midrule
\multicolumn{4}{l}{\textbf{Machine-learning corpus}} \\
\midrule
\texttt{2607.13468}~\cite{zang2026hive} & cs.CV & 19 & HIVE-3D: Hierarchical Voxel Enhancement for High-Quality 3D Scene Generation  \\
\texttt{2607.13475}~\cite{yang2026surgical} & cs.RO & 3 & Deformable State Estimation for Autonomous Surgical Tissue Retraction Under Partial Observability  \\
\texttt{2607.13511}~\cite{reddy2026externd} & cs.LG & 9 & ExTernD: Expanded-Rank Ternary Decomposition Ternary LLM PTQ with Accuracy Approaching Any Quantization Level  \\
\texttt{2607.13613}~\cite{tran2026fastcentnn} & cs.LG & 4 & FastCentNN: Accelerating Centroid Neural Network with Entropy Proxy  \\
\texttt{2607.13618}~\cite{deb2026stocktake} & cs.AI & 18 & STOCKTAKE: Measuring the Gap Between Perception and Action in LLM Agents with a Fair Oracle  \\
\texttt{2607.13749}~\cite{kam2026grokking} & cs.LG & 15 & Algebraic Representability as the Limiting Regime of Grokking: An Exactly Solvable Model with Holomorphic Activations  \\
\texttt{2607.14004}~\cite{wang2026agentoptimizers} & cs.AI & 16 & Do Agent Optimizers Compound? A Continual-Learning Evaluation on Terminal-Bench 2.0  \\
\texttt{2607.14008}~\cite{gursoy2026lighthouse} & cs.LG & 7 & Lighthouse RL: Sample-Efficient Circuit Optimization via Strategic Reset Points  \\
\texttt{2607.14024}~\cite{grillmeyer2026windsolar} & cs.LG & 17 & Improving Wind and Solar Power Prediction with Efficient Wrapper-based Feature Selection: An Empirical Study  \\
\texttt{2607.14176}~\cite{amati2026relay} & cs.LG & 6 & Low-Latency Relay Selection in NR-V2X Vehicular Communications via Graph Isomorphism Networks with Edge Features  \\
\texttt{2607.14287}~\cite{hasan2026xctsam} & cs.CV & 10 & XCT-SAM: Sequential Parameter-Efficient Domain Adaptation of SAM for Industrial XCT Defect Segmentation  \\
\texttt{2607.14293}~\cite{gaytan2026chargejumps} & quant-ph & 9 & Real-Time Detection of Charge Jumps in Superconducting Qubits with a Convolutional Neural Network  \\
\texttt{2607.14315}~\cite{makridis2026explainability} & cs.LG & 8 & Towards a Unified Multidimensional Explainability Metric: Evaluating Trustworthiness in AI Models  \\
\texttt{2607.14375}~\cite{soltanalian2026relu} & cs.CC & 16 & Random Parameter Noise Does Not Make Exact ReLU Verification Easy  \\
\texttt{2607.20554}~\cite{amati2026multihop} & cs.AI & 7 & AI-Driven Multi-Hop Relay Selection for Smart Urban NR-V2X Networks via Learning-to-Optimize Graph Neural Networks  \\
\bottomrule
\end{tabular}
\end{table}

\section{Error injection design and taxonomy} \label{app:taxonomy}

The injection pipeline has two pieces. First, a \emph{severity-balanced 3-copy schedule} fixes how the twelve (error type, severity) combinations are distributed across the three injected copies of each paper, so that every combination is chosen exactly once. Second, a \emph{combination registry} of 33 sub-types, distributed across those twelve (error type, severity) combinations, with two or three sub-types per combination, determines what each combination looks like in practice, with a verbatim mechanism block, an eligibility filter, and a fixed output format (a Pydantic schema) that the LLM-driven proposer must conform to.

\subsection{Severity-balanced 3-copy schedule} \label{app:design} Each paper features in four versions: a clean copy that carries the original LaTeX source, and three injected copies, each bearing one error per error type (four errors per copy). Across the three injected copies, every (error type, severity) combination is hit exactly once. A schematic schedule for one paper:
\begin{center}
\small
\begin{tabular}{l c c c c}
\toprule
& Citations & Numerics & Mathematics & General \\
\midrule
Copy 1 & $S_1$ & $S_2$ & $S_3$ & $S_1$ \\
Copy 2 & $S_2$ & $S_3$ & $S_1$ & $S_2$ \\
Copy 3 & $S_3$ & $S_1$ & $S_2$ & $S_3$ \\
\bottomrule
\end{tabular}
\end{center}
Reading down a column gives a permutation of $\{S_1, S_2, S_3\}$, so that each error type sees each severity exactly once across the three copies. The permutations are drawn independently for each error type, so the severities of the four errors in a copy are independent of one another.

\subsection{Injection error sub-types with verbatim examples}

Table~\ref{tab:severity} gives the operating definition of each (error type, severity) combination used for the astrophysics corpus, summarising the severity scale per error type from $S_1$ (subtle) through $S_2$ (clear) to $S_3$ (blatant). The injection registry refines this taxonomy into 33 sub-types distributed across the twelve combinations, with two or three sub-types per combination (e.g.\ Citations $\cdot$ $S_1$ admits both ``quantitative-qualifier mis-quote'' and ``methodological-qualifier drift''). On the ML corpus, the same registry was used with one substitution: since ML equations often carry no physical units, the mathematics $S_3$ sub-type `dimensional absurdity' was replaced by `shape or axis non-conformability' (a tensor-shape or index mismatch).

\begin{table}[h]
\caption{Severity levels per error type.} \label{tab:severity} \centering \small \renewcommand{\arraystretch}{1.15}
\begin{tabular}{p{1.6cm} p{3.4cm} p{3.6cm} p{3.6cm}}
\toprule
\textbf{Error type} & \textbf{$S_1$ (subtle)} & \textbf{$S_2$ (clear)} & \textbf{$S_3$ (blatant)} \\
\midrule
Citations &
quantity shifted by 1--2$\sigma$, or a significant
methodological qualifier dropped &
conditional / scope qualifier dropped, or methodology
category mis-attributed &
direction of the cited conclusion reversed, or topic
mis-attributed \\
Numerics &
small-magnitude inconsistency (abstract$\leftrightarrow$body
second-decimal drift, table totals slightly off) &
clearly outside tolerance (digit swap, percent-of
inconsistency, sample-size disagreement) &
structurally wrong (percentages summing $> 100\,\%$,
order-of-magnitude mismatch, blatantly wrong ratio) \\
Mathematics &
sign flip on a subleading term, coefficient slight
perturbation, or sum / integral bound shift &
term deletion, operator swap, or prefactor /
normalisation drop &
dimensional absurdity, wrong-but-plausible formula, or
patently false limit claim \\
General issues &
cosmetic / editorial (typo, rendering artefact, local
notation drift) & cross-reference broken,
Methods$\leftrightarrow$Results inconsistency,
phantom-evidence insertion &
abstract$\leftrightarrow$body contradiction,
sweeping unsupported generalisation, methodology vs
experiment contradiction \\
\bottomrule
\end{tabular}
\end{table}

For each sub-type to be injected into a paper, an LLM-driven proposer (Claude Sonnet 4.6 on the astrophysics corpus, Gemini 3.1 Pro on the ML corpus) is given the full paper LaTeX, a verbatim mechanism block, an eligibility filter, and a fixed output format (a Pydantic schema), and asked to return a single atomic edit. Each proposed edit then passes through a mechanical validator. The validator accepts an edit only if the text to be replaced occurs exactly once in a source file that is part of the compiled document, the replacement actually changes the text, and no citation key is added, removed, or altered; the edited copy must also compile.

Here we give one verbatim example per sub-type, all drawn from the astrophysics corpus, grouped by severity within each error type.

\subsubsection{Citations}

\paragraph{$S_1$ (subtle).}

\textbf{Quantitative-qualifier mis-quote.} \emph{Paper:} \texttt{2604.20203}. \emph{Original:} ``Here we adopt the result of \texttt{[cite]}, $P(\rm CJ, [Fe/H]>0) = 14.3^{+2.0}_{-1.8}\%$, as their definition of cold Jupiters in mass and semimajor axis is same to that in this work.''. \emph{Injected:} ``\dots $P(\rm CJ, [Fe/H]>0) = 16.5^{+2.0}_{-1.8}\%$, as their definition\dots''. The cited R21 paper reports $14.3^{+2.0}_{-1.8}\,\%$; the injection nudges the value upward by $+2.2\,\mathrm{pp}$ ($\sim 1.1\sigma$) while the cite key is preserved byte-for-byte.

\textbf{Methodological-qualifier drift.} \emph{Paper:} \texttt{2604.20975}. \emph{Original:} ``We model it using the Schechter-like parametric model introduced in \texttt{[cite]},''. \emph{Injected:} ``We model it using the parametric model introduced in \texttt{[cite]},''. The dropped functional-form qualifier (``Schechter-like'') is explicit in the cited paper and load-bearing for the model class.

\paragraph{$S_2$ (clear).}

\textbf{Conditional-claim drop / over-generalisation.} \emph{Paper:} \texttt{2604.20086}. \emph{Original:} ``stacked analyses using IceCube \texttt{[cite]} data have shown that neutrino searches from dSphs can provide competitive constraints on the dark matter annihilation cross section for several leptonic channels.''. \emph{Injected:} same sentence with the trailing qualifier ``for several leptonic channels'' dropped, generalising the cited result from leptonic channels to all annihilation channels.

\textbf{Methodology category mis-attribution.} \emph{Paper:} \texttt{2604.20080}. \emph{Original:} ``as observed in giant molecular clouds \texttt{[cite]}''. \emph{Injected:} ``as predicted by hydrodynamic simulations of giant molecular clouds \texttt{[cite]}''. Solomon et al.\ 1987 is purely observational; recasting the citing claim as ``predicted by hydrodynamic simulations'' mis-categorises the cited methodology.

\paragraph{$S_3$ (blatant).}

\textbf{Direction reversal.} \emph{Paper:} \texttt{2604.21099}. \emph{Original:} ``\texttt{[cite]} limit of $E\sim 5\times 10^{19}$ eV (or GZK cutoff) is the upper bound on the energy for cosmic-ray protons \dots''. \emph{Injected:} ``\dots is the lower bound on the energy for cosmic-ray protons \dots''. The polarity of the cited GZK conclusion is reversed (upper $\to$ lower bound).

\textbf{Topic mis-attribution.} \emph{Paper:} \texttt{2604.20344}. \emph{Original:} ``the conclusion reached previously for the Nuker model \texttt{[cite]}''. \emph{Injected:} ``the convergence criterion established for numerical Abel-transform inversions in triaxial stellar systems \texttt{[cite]}''. The reattributed topic (numerical Abel-transform inversions in triaxial systems) is unrelated to the cited paper's actual subject (the Nuker model).

\subsubsection{Numerics}

\paragraph{$S_1$ (subtle).}

\textbf{Abstract$\leftrightarrow$body second-decimal drift.} \emph{Paper:} \texttt{2604.20080}. \emph{Original (abstract):} ``The global CO(2--1)/CO(1--0) ratio is low ($R_{21}\!\approx\!0.29$--0.34)''. \emph{Injected (abstract):} ``\dots ($R_{21}\!\approx\!0.29$--0.35)''. The abstract's headline $R_{21}$ upper bound is perturbed in the second decimal; the body (Sect.\ 3.5 and Table 5) still reports $R_{21}=0.34$.

\textbf{Table-row total slight mismatch.} \emph{Paper:} \texttt{2604.20411}. \emph{Original (Table 1, $0.7\le z<1.0$ row, galaxy-count column):} \texttt{47279}. \emph{Injected:} \texttt{47269}. The row count is reduced by 10; the abstract still reports the unmodified mass-complete sample size of $\sim$\,290k.

\textbf{Repeated-constant slight drift.} \emph{Paper:} \texttt{2604.20971}. \emph{Original:} ``turbulence normalization $C$ ($=~360$ in the aforementioned HD study)''. \emph{Injected:} ``turbulence normalization $C$ ($=~370$ in the aforementioned HD study)''. The constant $C$ printed in Sec.\ 4.1 is bumped by $+10$; Sec.\ 3 (Eq.\ 6) and the Conclusion still print $C=360$.

\paragraph{$S_2$ (clear).}

\textbf{Headline-number digit swap.} \emph{Paper:} \texttt{2604.20142}. \emph{Original (abstract):} ``\dots remains centered around $\alpha \approx -1.5$''. \emph{Injected (abstract):} ``\dots remains centered around $\alpha \approx -5.1$''. A digit transposition $-1.5\to -5.1$ in the abstract; the body, table, and figure continue to report values centred near $-1.5$.

\textbf{Percent-of arithmetic inconsistency.} \emph{Paper:} \texttt{2604.20914}. \emph{Original:} ``$\sim 200$ times speedup, reducing the computation from about 13 hours to 4 minutes''. \emph{Injected:} ``\dots from about 13 hours to 6 minutes''. The same sentence advertises a $\sim 200\times$ speedup and a 13-hour baseline; $13\,\mathrm{h}/200=3.9\,\mathrm{min}$, so the printed runtime no longer matches the printed ratio.

\textbf{Integer-equal sample-size disagreement.} \emph{Paper:} \texttt{2604.20411}. \emph{Original (body):} ``$\sim$290k galaxies from the latest COSMOS2020 catalog''. \emph{Injected (body):} ``$\sim$29k galaxies from the latest COSMOS2020 catalog''. The abstract still says ``mass-complete sample of approximately 290,000 galaxies''---a factor-10 discrete disagreement between abstract and body.

\paragraph{$S_3$ (blatant).}

\textbf{Percentages do not sum to 100.} \emph{Paper:} \texttt{2604.20344}. \emph{Original:} ``A substantial fraction of the formal core--S\'ersic parameter space is therefore physically inadmissible.''. \emph{Injected:} ``A substantial fraction (approximately $40\,\%$) of the formal core--S\'ersic parameter space is therefore physically inadmissible, while the remaining $75\,\%$ corresponds to physically acceptable models.''. The split ``$40\,\%$ inadmissible $+\;75\,\%$ admissible'' totals $115\,\%$, structurally violating part-of-whole.

\textbf{Order-of-magnitude mismatch.} \emph{Paper:} \texttt{2604.20086}. \emph{Original (abstract):} ``in the Milky Way and in 14 dwarf spheroidal galaxies''. \emph{Injected (abstract):} ``in the Milky Way and in 140 dwarf spheroidal galaxies''. The body still states ``we analyzed 14 different dSphs'' and Table 1 lists 14 rows: a factor-10 mismatch between abstract and body.

\textbf{Ratio arithmetic blatantly wrong.} \emph{Paper:} \texttt{2604.20971}. \emph{Original:} ``symmetric mass ratio $q/(1+q)^2\approx q$ because $q\ll 1$''. \emph{Injected:} ``symmetric mass ratio $q/(1+q)^2\approx q/4$ because $q\ll 1$''. The leading-order approximation in the small-$q$ limit is replaced by a value four times smaller, numerically wrong by a factor of four for any small $q$.

\subsubsection{Mathematics}

\paragraph{$S_1$ (subtle).}

\textbf{Subleading-term sign flip.} \emph{Paper:} \texttt{2604.20344}. \emph{Original:} $I_b\,\exp\!\left[-b\,(R/R_e)^{1/m} + b\,(R_b/R_e)^{1/m}\right]$. \emph{Injected:} $I_b\,\exp\!\left[-b\,(R/R_e)^{1/m} - b\,(R_b/R_e)^{1/m}\right]$. The sign of the subleading $+b\,(R_b/R_e)^{1/m}$ term inside the core-S\'ersic sharp-transition exponential is flipped from $+$ to $-$, making the limiting form inconsistent with its general-form derivation elsewhere in the same paper.

\textbf{Coefficient slight perturbation.} \emph{Paper:} \texttt{2604.20086}. \emph{Original:} $\Omega^2(\phi)=1+2\xi M \kappa^2\phi$. \emph{Injected:} $\Omega^2(\phi)=1+4\xi M \kappa^2\phi$. The numerical prefactor on the linear term of the conformal factor doubles ($2\to 4$).

\textbf{Index / bound shift.} \emph{Paper:} \texttt{2604.20142}. \emph{Original:} $+\frac{1}{E}\sum_{n=1}^{\infty}\frac{(1+4E)^{-n}}{n^{2}}$. \emph{Injected:} $+\frac{1}{E}\sum_{n=0}^{\infty}\frac{(1+4E)^{-n}}{n^{2}}$. The lower bound of the Klein--Nishina kernel sum is shifted from $n=1$ to $n=0$, introducing a $1/0$ singularity at $n=0$ absent from the original Jones (1968) expansion.

\paragraph{$S_2$ (clear).}

\textbf{Term deletion.} \emph{Paper:} \texttt{2604.20080}. \emph{Original:} $\log\alpha_{\rm CO}=0.7-1.5\left[(12+\log{\rm O/H})-8.7\right]$. \emph{Injected:} $\log\alpha_{\rm CO}=0.7-1.5\left[(12+\log{\rm O/H})\right]$. The metallicity zero-point term ``$-8.7$'' is deleted from the Amorin+2016 $\log\alpha_{\rm CO}$ calibration; the formula remains valid \LaTeX{} but no longer reproduces the published values at the solar anchor.

\textbf{Operator swap.} \emph{Paper:} \texttt{2604.20262}. \emph{Original:} $S = \frac{1}{N}\sum_{(i,j)\in \mathcal{S}} V_{i,j}^{\rm cal}\, e^{\frac{2\pi i}{\lambda}\vec{n}\cdot(\vec{r}_j - \vec{r}_i)}$. \emph{Injected:} $S = \frac{1}{N}\prod_{(i,j)\in \mathcal{S}} V_{i,j}^{\rm cal}\, e^{\frac{2\pi i}{\lambda}\vec{n}\cdot(\vec{r}_j - \vec{r}_i)}$. The outer summation in the beamforming definition is swapped for a product, turning a coherent sum over baselines into a product of phased visibilities.

\textbf{Prefactor / normalisation drop.} \emph{Paper:} \texttt{2604.20411}. \emph{Original:} $\log\mathrm{SFR}=S_0-\log\left[(10^{M_\star}/10^{M_0})^{-\gamma_1}+(10^{M_\star}/10^{M_0})^{\gamma_2}\right]$. \emph{Injected:} $\log\mathrm{SFR}=S_0-\tfrac{1}{2}\log\left[(10^{M_\star}/10^{M_0})^{-\gamma_1}+(10^{M_\star}/10^{M_0})^{\gamma_2}\right]$. A spurious $1/2$ prefactor is inserted in front of the $\log$ on the RHS of the modified Lee+15 main-sequence parameterisation, breaking the stated reduction to the unmodified form when $\gamma_2=0$.

\paragraph{$S_3$ (blatant).}

\textbf{Dimensional absurdity.} \emph{Paper:} \texttt{2604.20411}. \emph{Original:} $\log\mathrm{SFR}=S_0-\log\left[(10^{M_\star}/10^{M_0})^{-\gamma_1}+(10^{M_\star}/10^{M_0})^{\gamma_2}\right]$. \emph{Injected:} $\log\mathrm{SFR}=S_0+10^{M_\star}-\log\left[(10^{M_\star}/10^{M_0})^{-\gamma_1}+(10^{M_\star}/10^{M_0})^{\gamma_2}\right]$. A bare additive term $10^{M_\star}$ (dimensions of $M_\odot$ since $M_\star$ is logarithmic) is inserted into a relation whose other terms are dimensionless.

\textbf{Wrong-but-plausible formula.} \emph{Paper:} \texttt{2604.20809}. \emph{Original:} $\gamma^2 = \frac{1}{1 - v^2}$. \emph{Injected:} $\gamma^2 = \frac{1}{\sqrt{1 - v^2}}$. The canonical Lorentz factor $\gamma^2$ is replaced by the structurally similar but incorrect $1/\!\sqrt{1-v^2}$.

\textbf{Patently false limit claim.} \emph{Paper:} \texttt{2604.20203}. \emph{Original:} $\frac{1}{B(a,b)} P^{a-1} (1-P)^{b-1}$. \emph{Injected:} $\frac{1}{B(a,b)} P^{a-1} (1-P)^{1-b}$. In the Beta-distribution PDF, the exponent $b-1$ is flipped to $1-b$, reversing the boundary behaviour at $P\to 1$ from the correct limit of $0$ (for $b>1$) to a divergence.

\subsubsection{General issues}

\paragraph{$S_1$ (subtle).}

\textbf{Typo / repeated word / homophone swap.} \emph{Paper:} \texttt{2604.20975}. \emph{Original:} ``thus have only a negligible effect on the detection efficiency''. \emph{Injected:} ``thus have only a negligible affect on the detection efficiency''. Classic homophone swap (``effect''$\to$``affect''); spell-checkers do not catch it.

\textbf{Rendering artefact.} \emph{Paper:} \texttt{2604.20086}. \emph{Original:} ``In section \texttt{\textbackslash ref\{sec:conclusions\}}, we provide our conclusions.''. \emph{Injected:} ``In section \texttt{\textbackslash ref\{sec:conclusion\}}, we provide our conclusions.''. The conclusions section has label \texttt{sec:conclusions}; the injected reference points at \texttt{sec:conclusion} (no trailing `s'), which compiles fine but renders as ``section ??'' in the PDF.

\textbf{Local notation drift.} \emph{Paper:} \texttt{2604.21099}. \emph{Original (Sec.\ 4 displayed equation):} $f_{MM}(E)=I_{MM}(E)/I_{v}(E)$. \emph{Injected:} $F_{MM}(E)=I_{MM}(E)/I_{v}(E)$. The surrounding paragraphs continue to call the same magnetic-monopole flux fraction $f_{MM}$.

\paragraph{$S_2$ (clear).}

\textbf{Broken cross-reference.} \emph{Paper:} \texttt{2604.21077}. \emph{Original:} ``\dots the region labeled \emph{Low-Energy SNe} in \texttt{Fig.~\textbackslash ref\{fig:boundsgag\}}''. \emph{Injected:} ``\dots in \texttt{Fig.~\textbackslash ref\{fig:CoolingBound\}}''. The reference target is shifted from the figure that actually contains the ``Low-Energy SNe'' region to a different (cooling-bound) figure; the cited figure no longer matches the surrounding claim.

\textbf{Methods$\leftrightarrow$Results inconsistency.} \emph{Paper:} \texttt{2604.20809}. \emph{Original:} ``where $\tau$ is the duration of the expanding phase.''. \emph{Injected:} ``where $\tau$ is the duration of the contracting phase.''. The same $\tau$ is identified as the duration of the \emph{expanding} phase elsewhere in Sec.\ 2 (and Sec.\ 3), so the local rewrite contradicts the surrounding text.

\textbf{Phantom-evidence insertion.} \emph{Paper:} \texttt{2604.20262}. \emph{Original:} ``To prevent cross-coupling between feeds, we selected those with $L > 1.8\,\mathrm{m}$.''. \emph{Injected:} ``\dots with $L > 1.8\,\mathrm{m}$ (shorter baselines showed visibly degraded phase stability in our preliminary checks).''. The parenthetical claim about ``visibly degraded phase stability in our preliminary checks'' has no corresponding entry in Results, the Appendix, or any figure.

\paragraph{$S_3$ (blatant).}

\textbf{Abstract$\leftrightarrow$body contradiction.} \emph{Paper:} \texttt{2604.20914}. \emph{Original (abstract):} ``With high fidelity, we achieve a $\sim\!200$-fold speedup for a simulated ringdown signal including six modes, providing a viable and scalable route for multi-mode ringdown analysis in the space context.''. \emph{Injected (abstract):} ``\dots however, the computational cost grows steeply with the number of modes, which is likely to restrict the approach to moderate mode counts in the space context.''. The body (Fig.\ 2 and surrounding text) explicitly states the opposite---the FIREFLY computational cost ``increases only mildly with $N$, demonstrating its scalability''.

\textbf{Sweeping unsupported generalisation.} \emph{Paper:} \texttt{2604.20971}. \emph{Original (Conclusion):} ``Hence, this work strongly motivates MHD simulations of gas disks embedded near-coalescence EMRIs to extract maximum science in the LISA era.''. \emph{Injected (Conclusion):} same sentence with an appended clause: ``\dots, and further demonstrates that turbulence-driven stochastic torques will generically dominate orbital evolution across all classes of compact binary inspirals embedded in any astrophysical accretion disk, regardless of the mass ratio or turbulence source.''. The added clause extrapolates from a single LISA EMRI / accretion-disk scenario to ``all classes of compact binary inspirals \dots regardless of the mass ratio or turbulence source''---unsupported by the paper's actual scope.

\textbf{Methodology vs experiment contradiction.} \emph{Paper:} \texttt{2604.20262}. \emph{Original:} ``The length of these baselines are all 3.2m.''. \emph{Injected:} ``The length of these baselines are all 3.2m. The visibilities from all 12 baselines are then jointly combined into a single stacked likelihood, which suppresses per-baseline systematics and ensures that the recovered N-S beam parameters are consistent across the full set of selected baselines.''. The Methods section is augmented with a clause asserting a ``stacked likelihood'' over all 12 baselines; the actual Results pipeline operates per-baseline rather than on a stacked likelihood.

\section{Panel-internal agreement} \label{app:panel-kappa}

Two LLM-judge panels feature in the adjudication pipeline (\S\ref{sec:reverif}): the detection-matching panel (Claude Sonnet 4.6, GPT-5.4, Gemini 3 Flash), which decides, for each injection, whether a given review detected it, and the general-issues false-positive panel (Claude Sonnet 4.6, GPT-5.4, Gemini 3.1 Pro), which classifies each general-issue flag raised on a clean copy as confirmed issue, false positive, or ambiguous. The rates in the main text rest on these panels, so we report here how well the judges within each panel agree with one another, on both corpora. For the matching panel we use the first-round votes only, so that the PDF-attached escalation pass is not mixed in.

\begin{figure}[h]
\centering \includegraphics[width=0.9\linewidth]{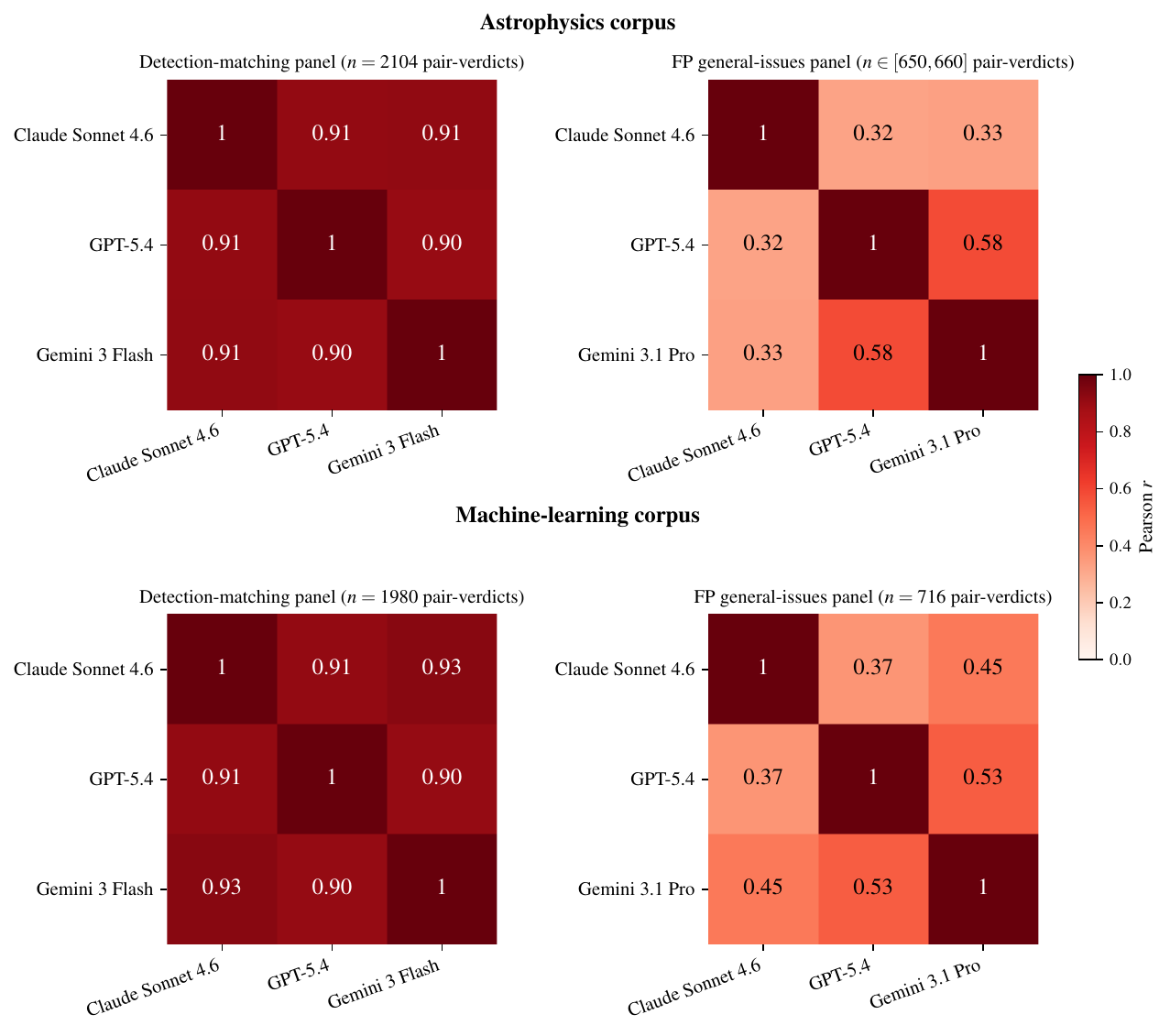} \caption{Pearson $r$ between judges within each adjudication panel, on binary votes, for the astrophysics corpus (top) and the ML corpus (bottom). Left: detection-matching panel, first-round votes. Right: general-issues false-positive panel. $n$ is the number of pair-verdicts.} \label{fig:corr-panel-app}
\end{figure}

Figure~\ref{fig:corr-panel-app} shows the pairwise Pearson $r$ between judges on their binary votes. The matching panel agrees closely, $r = 0.90$--$0.91$ on astrophysics ($n = 2{,}104$ pair-verdicts) and $0.90$--$0.93$ on ML ($n = 1{,}980$): the task is a lookup against an enumerated injection record, and high agreement is what one expects when an objective key exists. The false-positive panel agrees far less, $r = 0.32$--$0.58$ on astrophysics ($n = 650$--$660$) and $0.37$--$0.53$ on ML ($n = 716$): whether a flagged point is a real but minor issue or a spurious one is ambiguous when no key is available, which is why this task is given to a three-judge panel that sees the full PDF. The two corpora differ little in this respect: a paper-level bootstrap puts every ML--astrophysics difference within $\pm 0.12$, and none of the $95\,\%$ intervals excludes zero. We also computed Cohen's $\kappa$ for every pair. It never differs from $r$ by more than $0.01$ on the matching panel or $0.03$ on the false-positive panel, so the choice of agreement measure does not affect these conclusions.

\section{Adjudication-panel family-pairing test} \label{app:family-pairing}

Several reviewer systems share an LLM provider family with one of the three judges of the detection-matching panel. Following~\cite{panickssery2024judge}, this raises a self-preference concern: a judge might vote ``detected'' more readily on output from its own family than a judge from another family would on the same injection, which would inflate the detection rates of those reviewer systems. We test this for both corpora on the first-round panel votes themselves, comparing the judges within each injection, rather than by re-adjudicating with a substitute panel.

To define the test, let $\mathrm{A}$, $\mathrm{O}$, $\mathrm{G}$ denote the Anthropic (Sonnet 4.6), OpenAI (GPT-5.4), and Google (Gemini 3 Flash) judges. For each of the eight reviewer systems built on a single LLM family (the six single-LLM configurations and the two Claude Code configurations), let $r^{(s)}_f \in [0,1]$ denote the first-round detection rate of judge $f$ on system $s$'s injections. With $\mathrm{self}(s)$ the family of system $s$ and $\mathrm{disj}(s) = \{\mathrm{A}, \mathrm{O}, \mathrm{G}\} \setminus \{\mathrm{self}(s)\}$,
$$\Delta_s \;=\; r^{(s)}_{\mathrm{self}(s)} \;-\; \tfrac{1}{2}\sum_{f \in \mathrm{disj}(s)} r^{(s)}_f .$$
A positive $\Delta_s$ means that the same-family judge detects more often than the other two. A judge that is more generous overall, voting ``detected'' more often on every system's output, would produce a positive $\Delta_s$ for all systems of its own family without any self-preference. To separate the two, we also report an adjusted version $\widetilde\Delta_s$, in which each judge's rate on a system is measured relative to that judge's rate pooled over all reviewer systems. These pooled rates are given in Table~\ref{tab:fampair-pooled}, which shows that the three judges are close to one another, within about $3\,\mathrm{pp}$ on both corpora, with OpenAI slightly the most generous. \skepthical{} and the external reviewers are excluded from $\Delta_s$, since \skepthical{} uses agents from all three judge families and the external reviewers either cannot be attributed to one family (Stanford) or share a family with no judge (open-weights). We note that all CIs in this appendix are $95\,\%$ paper-level bootstrap intervals.

\begin{table}[h]
\centering \footnotesize
\caption{Pooled first-round detection rate by judge, across all reviewer systems' detection-matching injections ($2{,}104$ on astrophysics, $1{,}980$ on ML). $95\,\%$ paper-level bootstrap CIs.}
\label{tab:fampair-pooled}
\begin{tabular}{lcc}
\toprule
Judge & Astrophysics & ML \\
\midrule
Anthropic (Sonnet 4.6)  & $36.5\,\%\;[33.1, 40.1]$ & $44.2\,\%\;[41.0, 47.4]$ \\
OpenAI (GPT-5.4)        & $38.3\,\%\;[34.8, 41.8]$ & $46.4\,\%\;[43.3, 49.5]$ \\
Google (Gemini 3 Flash) & $35.2\,\%\;[31.8, 38.9]$ & $43.1\,\%\;[39.4, 46.8]$ \\
\bottomrule
\end{tabular}
\end{table}

Table~\ref{tab:fampair-perlab} reports $\Delta_s$ and $\widetilde\Delta_s$ for each of the eight systems on each corpus. With under $200$ injections per system, the interval of a single system is wide, and most of them include zero, so no single row can establish or rule out self-preference on its own. The test therefore rests on the mean of $\Delta_s$ over the eight systems, whose bootstrap interval is given in the last row of each block.

\begin{table}[h]
\centering \footnotesize
\caption{Difference between the same-family judge's detection rate and the mean of the other two judges' rates, within injections, for each single-family reviewer system on each corpus. ``self'' is the system's LLM family ($\mathrm{A}$ Anthropic, $\mathrm{O}$ OpenAI, $\mathrm{G}$ Google). $\Delta$ is defined in the text; $\widetilde\Delta$ centres each judge's per-system rate on that judge's pooled rate. Positive values would indicate self-preference. $95\,\%$ paper-level bootstrap CIs in brackets; the mean row resamples papers, recomputes every $\Delta_s$, and averages.}
\label{tab:fampair-perlab}
\begin{tabular}{llrrr}
\toprule
Reviewer system & self & $\Delta$ (raw, pp) & $\widetilde\Delta$ (adj., pp) & $n$ \\
\midrule
\multicolumn{5}{l}{\textbf{Astrophysics corpus}} \\
\midrule
Claude Opus 4.7 (free-form)              & $\mathrm{A}$ & $-2.96\;[-5.83, +0.00]$ & $-2.72\;[-5.34, -0.12]$ & $169$ \\
GPT-5.5 (free-form)                      & $\mathrm{O}$ & $+1.41\;[-0.56, +3.45]$ & $-0.99\;[-2.54, +0.49]$ & $177$ \\
Gemini 3.1 Pro (free-form)               & $\mathrm{G}$ & $-3.95\;[-6.94, -1.41]$ & $-1.79\;[-4.18, +0.26]$ & $177$ \\
Claude Opus 4.7 (structured)             & $\mathrm{A}$ & $-2.07\;[-5.94, +1.96]$ & $-1.83\;[-5.36, +1.80]$ & $169$ \\
GPT-5.5 (structured)                     & $\mathrm{O}$ & $+0.00\;[-2.03, +2.19]$ & $-2.40\;[-4.30, -0.29]$ & $173$ \\
Gemini 3.1 Pro (structured)              & $\mathrm{G}$ & $-3.11\;[-6.50, -0.28]$ & $-0.94\;[-4.17, +1.84]$ & $177$ \\
Claude Code (structured)                 & $\mathrm{A}$ & $-1.41\;[-3.67, +0.57]$ & $-1.17\;[-3.55, +1.11]$ & $177$ \\
Claude Code (free-form)                  & $\mathrm{A}$ & $+1.13\;[-1.11, +3.11]$ & $+1.37\;[-0.86, +3.50]$ & $177$ \\
Mean across the eight systems            &              & $-1.37\;[-2.41, -0.39]$ & $-1.31\;[-2.36, -0.34]$ &       \\
\midrule
\multicolumn{5}{l}{\textbf{Machine-learning corpus}} \\
\midrule
Claude Opus 4.7 (free-form)              & $\mathrm{A}$ & $-1.94\;[-3.89, +0.28]$ & $-1.44\;[-3.56, +0.96]$ & $180$ \\
GPT-5.5 (free-form)                      & $\mathrm{O}$ & $+1.94\;[-0.28, +4.17]$ & $-0.73\;[-3.13, +1.64]$ & $180$ \\
Gemini 3.1 Pro (free-form)               & $\mathrm{G}$ & $-1.94\;[-4.17, +0.00]$ & $+0.23\;[-1.79, +2.12]$ & $180$ \\
Claude Opus 4.7 (structured)             & $\mathrm{A}$ & $+0.56\;[-1.94, +3.61]$ & $+1.06\;[-1.21, +3.84]$ & $180$ \\
GPT-5.5 (structured)                     & $\mathrm{O}$ & $+0.83\;[-0.56, +2.22]$ & $-1.84\;[-3.18, -0.45]$ & $180$ \\
Gemini 3.1 Pro (structured)              & $\mathrm{G}$ & $+0.28\;[-1.67, +2.50]$ & $+2.45\;[+0.86, +4.37]$ & $180$ \\
Claude Code (structured)                 & $\mathrm{A}$ & $-1.11\;[-3.61, +1.39]$ & $-0.61\;[-3.13, +1.79]$ & $180$ \\
Claude Code (free-form)                  & $\mathrm{A}$ & $+0.00\;[-4.72, +4.72]$ & $+0.51\;[-3.79, +5.08]$ & $180$ \\
Mean across the eight systems            &              & $-0.17\;[-0.97, +0.66]$ & $-0.05\;[-0.73, +0.65]$ &       \\
\bottomrule
\end{tabular}
\end{table}

This mean is $-1.37\,\mathrm{pp}$ ($95\,\%$ CI $[-2.41, -0.39]$) on astrophysics and $-0.17\,\mathrm{pp}$ ($[-0.97, +0.66]$) on ML; with the adjustment it is $-1.31$ ($[-2.36, -0.34]$) and $-0.05$ ($[-0.73, +0.65]$). Self-preference would make these means positive. They are not: on ML the mean is consistent with zero, on astrophysics it is slightly negative, and in no case does the interval reach beyond $+0.7\,\mathrm{pp}$. On both corpora, then, judges do not favour reviewers of their own family by any amount that could inflate a detection rate. The slightly negative astrophysics value should not be read as judges disfavouring their own family either: it is about $1\,\mathrm{pp}$, only two of the eight per-system intervals exclude zero, and effects unrelated to family (e.g., differences in how thoroughly each judge reads, judge style priors, the randomness of a single panel run) could produce a difference of that size.

The split panels allow for a second test. On the $2$-$1$ splits, the injections on which one judge disagreed with the other two, we ask how often the lone dissenter is the judge of the reviewer's own family. If every judge dissents equally often, that share is one third. Self-preference would push it below one third on the $2$-$1$ \emph{detected} panels, where two judges found the error and the same-family judge would be expected to be among them, and above one third on the $1$-$2$ \emph{not-detected} panels, where only one judge found it and self-preference would make that one the same-family judge. Table~\ref{tab:fampair-dissent} shows that the same-family judge is the lone dissenter in $25.2\,\%$ of the $107$ astrophysics splits and $23.2\,\%$ of the $108$ ML splits, and that on the not-detected panels, where self-preference would raise the share above one third, it is $16.9\,\%$ on astrophysics and $24.3\,\%$ on ML. Neither corpus shows the pattern that self-preference would produce.

\begin{table}[h]
\centering \footnotesize
\caption{Share of $2$-$1$ split panels in which the same-family judge is the lone dissenter, pooled across the eight single-family reviewer systems. Null $= 33.\overline 3\,\%$. $95\,\%$ paper-level bootstrap CIs.}
\label{tab:fampair-dissent}
\begin{tabular}{lcrcr}
\toprule
 & \multicolumn{2}{c}{Astrophysics} & \multicolumn{2}{c}{ML} \\
Quantity & Same-family share & $n$ & Same-family share & $n$ \\
\midrule
All $2$-$1$ splits                 & $25.2\,\%\;[20.2, 29.9]$ & $107$ & $23.2\,\%\;[16.1, 30.9]$ & $108$ \\
$2$-$1$ \emph{detected} splits     & $38.1\,\%\;[23.9, 51.4]$ & $42$  & $21.1\,\%\;[10.5, 33.3]$ & $38$ \\
$1$-$2$ \emph{not-detected} splits & $16.9\,\%\;[11.1, 23.1]$ & $65$  & $24.3\,\%\;[14.5, 33.8]$ & $70$ \\
\bottomrule
\end{tabular}
\end{table}

\skepthical{} cannot be tested in this way, because its reviews are written by agents from all three families, so every judge is a same-family judge and there is none left to compare against. The question for \skepthical{} is nonetheless answered indirectly. If a judge favoured reviews written by models of its own family, this would appear as a positive $\Delta_s$ for the systems built on that family: the four Anthropic-based systems for the Anthropic judge, and two systems each for the OpenAI and Google judges. It appears for none of the three judges, on either corpus, so there is no sign of the mechanism that could favour \skepthical{}.

Two related concerns remain outside this test. First, the test can only detect a preference specific to one judge. A preference that is shared by all three, for instance for a certain style of review prose, would affect every reviewer system alike and cannot be detected by this test. Second, the human ground truth against which the panel can be checked is limited to the one-paper check of Appendix~\ref{app:human}.

Finally, as an additional test, we ask how much the results depend on the size of the panel. In particular, we recompute every system's detection rate with one judge family removed from the first-round panel, taking the majority of the remaining two votes and resolving $1$-$1$ ties either strictly (not detected) or leniently (detected). On astrophysics, the same three systems stay on top whichever judge is removed, and with strict ties their order is also unchanged. In the worst case for \skepthical{} (Google removed, strict ties) its rate falls from $70.6$ to $63.3\,\%$, a change of $7.3\,\mathrm{pp}$ (CI $[3.9, 11.5]$), and its lead over structured GPT-5.5 shrinks from $9.3$ to $2.6\,\mathrm{pp}$. On ML, structured GPT-5.5 stays in the first position whichever judge is removed, \skepthical{} and structured Claude Code, $3.3\,\mathrm{pp}$ apart at baseline, swap second and third place in two of the six variants, and \skepthical{}'s largest change is $5.6\,\mathrm{pp}$ (CI $[2.2, 10.0]$). With lenient ties, \skepthical{}'s rate stays within $1.7\,\mathrm{pp}$ of its baseline in every variant on both corpora. Per-system rates under each variant are released with the benchmark.

The script that produces every result in this appendix is released with the benchmark; it makes no LLM calls and runs in about a minute.

\section{Human adjudication} \label{app:human}

The detection rates in this paper are computed from the verdicts of an LLM panel. As a validation of these verdicts, one of the authors annotated every detection decision for one astrophysics paper, without knowledge of the panel's verdicts or of which system had written each review.

The paper, \texttt{2604.20975}, was drawn at random from the 15-paper corpus. Its three injected copies carry four injections each, and each copy was reviewed by the 12 systems, giving 36 reviews and 144 adjudications, each on whether a given review detected a given injection. Twelve of the 144 decisions concern three reviews for which no panel verdict was recorded, leaving 132 comparisons.

On these 132, the annotator and the panel agree on 127 ($96.2\,\%$; Cohen's $\kappa = 0.914$). The five disagreements are all in the same direction: the panel credited a detection that the annotator rejected. The panel's detection rate on this paper is therefore slightly higher than the annotator's, $34.8$ against $31.1\,\%$.

Three limitations apply. First, the check covers a single astrophysics paper. Annotating every review of a paper against every injection is labour-intensive, and extending it to further papers or to the ML corpus was beyond the resources of this work, so the agreement reported here is evidence about the panel on one paper rather than a corpus-wide validation. Second, with a single annotator, inter-annotator agreement cannot be measured, so a disagreement with the panel cannot be attributed to the panel rather than to the annotator. Third, the check covers detection only: the false-positive panel, which Appendix~\ref{app:panel-kappa} shows to be the less reliable of the two, was not checked against a human.

\section{Significance tests: configuration comparisons and the severity trend} \label{app:stats}

In this appendix we present the results of the tests behind two kinds of statements in \S\ref{sec:res-detection}: that a given reviewer configuration detects more than another one, and that detection rises with injected severity. Both use only the per-injection verdicts and severity labels already computed, resampling the 15 papers of a corpus ($n_\mathrm{boot} = 10{,}000$, against $2000$ for the intervals of \S\ref{sec:res-detection}) or permuting severity labels within papers; no new LLM calls are made.

\paragraph{Comparisons between configurations.} For each comparison made in \S\ref{sec:res-detection}, we resample the 15 papers with replacement, recompute the two rates on the same resampled papers, and take their difference. Table~\ref{tab:stats-pairs} gives the difference on the full corpus and the fraction of resamples in which its sign reverses, for the total rate and for each error type. We recall that the main text calls a difference significant when this fraction is below $0.05$. Four readings of the table are used there. First, the structured prompt raises the total rate significantly for Claude Opus 4.7 and Gemini 3.1 Pro on astrophysics, for all three providers on ML, and within Claude Code on both corpora; GPT-5.5 on astrophysics is the one case where it does not. Second, giving the model tools helps only with the structured prompt and only on astrophysics, where the gain is significant on numerics and mathematics; on ML no change is significant, and with the free-form prompt Claude Code detects significantly less than the single call on both corpora. Third, \skepthical{} is significantly ahead of structured Claude Code on citations and general issues on astrophysics and significantly behind it on numerics, with no significant difference on ML. Fourth, against structured GPT-5.5, \skepthical{} is significantly ahead on astrophysics on the total rate and on citations, and not significantly different on ML in any column.

\begin{table}[h]
\centering \scriptsize \setlength{\tabcolsep}{3.5pt}
\caption{Difference in detection rate (pp) for each comparison made in \S\ref{sec:res-detection}, with the paired paper-level bootstrap probability that its sign reverses in parentheses ($n_\mathrm{boot} = 10{,}000$; ``$<\!0.001$'' means fewer than one resample in a thousand reversed it). Bold: probability below $0.05$. ``Single call'' is the single-LLM configuration with the same model (Claude Opus 4.7) and prompt as the Claude Code configuration.}
\label{tab:stats-pairs}
\resizebox{\textwidth}{!}{%
\begin{tabular}{lrrrrr}
\toprule
Comparison & Total & Citations & Numerics & Mathematics & General \\
\midrule
\multicolumn{6}{l}{\textbf{Astrophysics corpus}} \\
\midrule
Structured vs free-form, Claude Opus 4.7     & $\mathbf{+16.6}$ ($<\!0.001$) & $+11.6$ ($0.060$) & $\mathbf{+23.3}$ ($0.004$) & $\mathbf{+22.5}$ ($<\!0.001$) & $+9.3$ ($0.135$) \\
Structured vs free-form, GPT-5.5             & $+0.8$ ($0.388$) & $-4.0$ ($0.322$) & $-0.5$ ($0.464$) & $+1.8$ ($0.410$) & $+6.1$ ($0.117$) \\
Structured vs free-form, Gemini 3.1 Pro      & $\mathbf{+25.4}$ ($<\!0.001$) & $\mathbf{+6.7}$ ($0.034$) & $\mathbf{+44.4}$ ($<\!0.001$) & $\mathbf{+42.9}$ ($<\!0.001$) & $+8.9$ ($0.161$) \\
Structured vs free-form, Claude Code         & $\mathbf{+32.2}$ ($<\!0.001$) & $\mathbf{+17.8}$ ($0.014$) & $\mathbf{+48.9}$ ($<\!0.001$) & $\mathbf{+52.4}$ ($<\!0.001$) & $+11.1$ ($0.171$) \\
Claude Code vs single call, free-form        & $\mathbf{-7.9}$ ($0.016$) & $-0.4$ ($0.418$) & $\mathbf{-13.6}$ ($0.032$) & $-13.5$ ($0.082$) & $-4.4$ ($0.109$) \\
Claude Code vs single call, structured       & $\mathbf{+7.8}$ ($0.031$) & $+5.7$ ($0.124$) & $\mathbf{+12.0}$ ($0.037$) & $\mathbf{+16.4}$ ($0.041$) & $-2.6$ ($0.358$) \\
\skepthical{} vs Claude Code, structured     & $\mathbf{+9.6}$ ($<\!0.001$) & $\mathbf{+20.0}$ ($0.033$) & $\mathbf{-11.1}$ ($0.043$) & $+2.4$ ($0.440$) & $\mathbf{+26.7}$ ($0.007$) \\
\skepthical{} vs GPT-5.5, structured         & $\mathbf{+9.3}$ ($0.001$) & $\mathbf{+28.5}$ ($<\!0.001$) & $+0.5$ ($0.475$) & $-1.8$ ($0.436$) & $+9.5$ ($0.105$) \\
\midrule
\multicolumn{6}{l}{\textbf{Machine-learning corpus}} \\
\midrule
Structured vs free-form, Claude Opus 4.7     & $\mathbf{+28.3}$ ($<\!0.001$) & $\mathbf{+44.4}$ ($<\!0.001$) & $\mathbf{+28.9}$ ($<\!0.001$) & $\mathbf{+37.8}$ ($<\!0.001$) & $+2.2$ ($0.425$) \\
Structured vs free-form, GPT-5.5             & $\mathbf{+17.2}$ ($<\!0.001$) & $\mathbf{+28.9}$ ($0.014$) & $\mathbf{+11.1}$ ($0.026$) & $\mathbf{+13.3}$ ($<\!0.001$) & $\mathbf{+15.6}$ ($0.008$) \\
Structured vs free-form, Gemini 3.1 Pro      & $\mathbf{+40.0}$ ($<\!0.001$) & $\mathbf{+26.7}$ ($<\!0.001$) & $\mathbf{+53.3}$ ($<\!0.001$) & $\mathbf{+51.1}$ ($<\!0.001$) & $\mathbf{+28.9}$ ($<\!0.001$) \\
Structured vs free-form, Claude Code         & $\mathbf{+38.9}$ ($<\!0.001$) & $\mathbf{+44.4}$ ($<\!0.001$) & $\mathbf{+44.4}$ ($<\!0.001$) & $\mathbf{+46.7}$ ($<\!0.001$) & $\mathbf{+20.0}$ ($0.002$) \\
Claude Code vs single call, free-form        & $\mathbf{-9.4}$ ($0.010$) & $+2.2$ ($0.423$) & $-13.3$ ($0.091$) & $-15.6$ ($0.082$) & $\mathbf{-11.1}$ ($0.026$) \\
Claude Code vs single call, structured       & $+1.1$ ($0.425$) & $+2.2$ ($0.436$) & $+2.2$ ($0.459$) & $-6.7$ ($0.287$) & $+6.7$ ($0.243$) \\
\skepthical{} vs Claude Code, structured     & $+3.3$ ($0.268$) & $-6.7$ ($0.222$) & $+4.4$ ($0.363$) & $+6.7$ ($0.312$) & $+8.9$ ($0.142$) \\
\skepthical{} vs GPT-5.5, structured         & $-4.4$ ($0.107$) & $-4.4$ ($0.317$) & $+0.0$ ($0.500$) & $-13.3$ ($0.053$) & $+0.0$ ($0.500$) \\
\bottomrule
\end{tabular}}
\end{table}

\paragraph{Severity trend.} The question is whether detection rises from $S_1$ to $S_3$. For each error type, we pool the detection counts of the reviewers with non-negligible detection (all but OpenReviewer-8B and PaperAudit-Bench Deep), so that each paper contributes one rate per severity level rather than a binary outcome per reviewer, and the number of tests falls from one per reviewer and error type (48 on astrophysics, 44 on ML) to four. We then test for an upward trend with Page's $L$ statistic (mid-ranks for ties), obtain the $p$-value by permuting the severity labels within papers, and apply the Benjamini--Hochberg correction at $\alpha = 0.05$ across the four tests (Table~\ref{tab:stats-monotone}). On astrophysics, the upward trend is significant for citations only; for numerics it falls just short (raw $p = 0.026$ against a corrected threshold of $0.025$), and for mathematics and general issues the rate at $S_2$ is lower than at $S_1$, so there is no monotone trend to detect. On ML, the trend is significant for citations, numerics, and mathematics, and general issues again show a lower rate at $S_2$ than at $S_1$ ($71$, $45$, and $81\,\%$ at $S_1$, $S_2$, $S_3$). The lower $S_2$ rate for general issues has a simple cause: the $S_1$ injections of that type (a typo or a repeated word) are easier to spot than the $S_2$ ones (a Methods--Results inconsistency or phantom evidence). Testing each reviewer and error type separately on the binary outcomes (same correction) rejects one combination on astrophysics, free-form GPT-5.5 on numerics ($42 \to 100\,\%$), and two on ML, the same combination ($47 \to 100\,\%$) and structured Claude Opus 4.7 on citations ($27 \to 80\,\%$); with $9$--$15$ papers per combination and a correction over all of them these tests are too conservative to resolve trends, and the pooled tests are the ones to read.

\begin{table}[h]
\centering \footnotesize
\caption{Severity trend, pooled over the reviewers with non-negligible detection (ten on astrophysics, nine on ML). Rates are the per-(paper, severity) averages over the pooled reviewers; $p_\mathrm{raw}$ is the within-paper severity-permutation $p$-value; the last column marks the tests that pass the Benjamini--Hochberg correction at $\alpha = 0.05$ across the four error types of a corpus.}
\label{tab:stats-monotone}
\begin{tabular}{lrrrrc}
\toprule
Error type  & $S_1$ & $S_2$ & $S_3$ & $p_\mathrm{raw}$ & BH \\
\midrule
\multicolumn{6}{l}{\textbf{Astrophysics corpus}} \\
\midrule
Citations   & $8.7\,\%$  & $13.1\,\%$ & $24.3\,\%$ & $0.005$ & \checkmark \\
Numerics    & $53.6\,\%$ & $58.7\,\%$ & $72.0\,\%$ & $0.026$ & \\
Mathematics & $47.3\,\%$ & $40.4\,\%$ & $59.3\,\%$ & $0.126$ & \\
General     & $53.5\,\%$ & $41.0\,\%$ & $69.3\,\%$ & $0.397$ & \\
\midrule
\multicolumn{6}{l}{\textbf{Machine-learning corpus}} \\
\midrule
Citations   & $21.5\,\%$ & $31.1\,\%$ & $51.9\,\%$ & $0.006$ & \checkmark \\
Numerics    & $57.0\,\%$ & $69.6\,\%$ & $71.1\,\%$ & $0.030$ & \checkmark \\
Mathematics & $54.8\,\%$ & $58.5\,\%$ & $64.4\,\%$ & $0.029$ & \checkmark \\
General     & $71.1\,\%$ & $45.2\,\%$ & $80.7\,\%$ & $0.143$ & \\
\bottomrule
\end{tabular}
\end{table}

The scripts to reproduce the results in this appendix are released with the benchmark and run in under a minute on a single CPU.

\section{Structured-concise prompt baseline} \label{app:concise}

The structured single-LLM configuration of \S\ref{sec:systems} uses \skepthical{}'s full operating instructions as its prompt. Its gain over the free-form prompt can therefore come from two sources that the main comparison cannot separate: simply telling the model which four kinds of error to look for, and more detailed instructions on how to look for them. To separate the two, we wrote a 471-word `concise structured' prompt that does only the first. This prompt asks the reviewer to examine citation support, numerical consistency, mathematical correctness, and logical or cross-section consistency, with a brief example of each. We ran it with the three single-LLM providers and inside Claude Code on all 60 copies of each corpus.

Table~\ref{tab:concise} shows the results next to those of the free-form and structured prompts. The concise prompt captures most of the detection gain of the structured one. Over the free-form prompt it raises total detection for every provider, by $8$--$16\,\mathrm{pp}$ on astrophysics and $12$--$38\,\mathrm{pp}$ on ML (all significant except Gemini on astrophysics, which falls just short), and inside Claude Code it accounts for $24.3$ of the $32.2\,\mathrm{pp}$ between the free-form and structured prompts on astrophysics and $32.3$ of $38.9\,\mathrm{pp}$ on ML. For the single LLMs, the structured prompt adds little over the concise one, and not uniformly: $+0.8$, $-7.6$, and $+16.4\,\mathrm{pp}$ for Claude, GPT-5.5, and Gemini on astrophysics, $+6.3$, $+5.0$, and $+2.2\,\mathrm{pp}$ on ML. Averaged over the three providers, the only significant per-type gain of the structured prompt on astrophysics is on general issues ($+14.0\,\mathrm{pp}$), whereas on ML it gains on citations ($+16.0$) and general issues ($+12.8$) and loses on mathematics ($-8.8$). Inside Claude Code, by contrast, the step from the concise to the structured prompt is significant on both corpora ($+7.9$ and $+6.6\,\mathrm{pp}$) and concentrated in citations and general issues.

The strongest concise control is GPT-5.5, and \skepthical{}'s total detection is not significantly different from it: $70.6$ against $68.9\,\%$ on astrophysics ($+1.7\,\mathrm{pp}$) and $76.7$ against $76.1\,\%$ on ML ($+0.6\,\mathrm{pp}$). The margin reported in \S\ref{sec:res-detection} against the structured GPT-5.5 therefore depends on the comparison prompt. Per error type, \skepthical{} remains significantly ahead of concise GPT-5.5 on citations and general issues on astrophysics and significantly behind it on numerics and mathematics. The two corpora also differ in how the prompts affect the spread between providers: on ML the concise prompt already reduces the spread in total detection, from $52.8$ to $27.2\,\mathrm{pp}$ ($30.0$ with the structured prompt), whereas on astrophysics only the structured prompt does ($48.6$, $48.0$, and $24.0\,\mathrm{pp}$ for the three prompts).

On astrophysics, the concise prompt's gain in detection comes with many more false positives. Its clean-copy false-positive rate is $17$--$27\,\%$ for the single LLMs, against $1$--$9\,\%$ with the free-form prompt and $3$--$6\,\%$ with the structured one, and $16.7\,\%$ inside Claude Code against $7.4$ and $3.5\,\%$. The excess consists mostly of mathematics flags. On ML, on the other hand, the two structured prompts have similar false-positive rates ($11$--$21\,\%$ concise structured, $11$--$23\,\%$ structured).

\begin{table}[h]
\centering \footnotesize \setlength{\tabcolsep}{4pt}
\caption{Detection rate (\%) by error type and clean-copy false-positive rate (\%) for the free-form (27 words), concise structured (471 words), and structured (5,946 words) prompts, for the three single-LLM providers and inside Claude Code, on each corpus. Rates and conventions as in Table~\ref{tab:headline} and Figure~\ref{fig:fp}; paper-level bootstrap half-widths are of the same size as in Table~\ref{tab:headline} and are omitted.}
\label{tab:concise}
\begin{tabular}{llrrrrrr}
\toprule
Model & Prompt & Total & Citations & Numerics & Mathematics & General & FP \\
\midrule
\multicolumn{8}{l}{\textbf{Astrophysics corpus}} \\
\midrule
Claude Opus 4.7  & free-form & $36.7$ & $9.3$ & $55.8$ & $32.5$ & $48.8$ & $1.0$ \\
                 & concise structured & $52.5$ & $26.7$ & $86.7$ & $52.4$ & $44.4$ & $19.3$ \\
                 & structured & $53.3$ & $20.9$ & $79.1$ & $55.0$ & $58.1$ & $5.2$ \\
GPT-5.5          & free-form & $60.5$ & $22.2$ & $80.0$ & $73.8$ & $66.7$ & $8.7$ \\
                 & concise structured & $68.9$ & $28.9$ & $88.9$ & $90.5$ & $68.9$ & $17.1$ \\
                 & structured & $61.3$ & $18.2$ & $79.5$ & $75.6$ & $72.7$ & $3.4$ \\
Gemini 3.1 Pro   & free-form & $11.9$ & $4.4$ & $13.3$ & $2.4$ & $26.7$ & $7.0$ \\
                 & concise structured & $20.9$ & $4.4$ & $28.9$ & $40.5$ & $11.1$ & $27.3$ \\
                 & structured & $37.3$ & $11.1$ & $57.8$ & $45.2$ & $35.6$ & $5.6$ \\
Claude Code      & free-form & $28.8$ & $8.9$ & $42.2$ & $19.0$ & $44.4$ & $7.4$ \\
                 & concise structured & $53.1$ & $20.0$ & $88.9$ & $71.4$ & $33.3$ & $16.7$ \\
                 & structured & $61.0$ & $26.7$ & $91.1$ & $71.4$ & $55.6$ & $3.5$ \\
\midrule
\multicolumn{8}{l}{\textbf{Machine-learning corpus}} \\
\midrule
Claude Opus 4.7  & free-form & $43.9$ & $13.3$ & $55.6$ & $40.0$ & $66.7$ & $5.9$ \\
                 & concise structured & $65.9$ & $36.4$ & $79.5$ & $86.4$ & $61.4$ & $10.9$ \\
                 & structured & $72.2$ & $57.8$ & $84.4$ & $77.8$ & $68.9$ & $11.5$ \\
GPT-5.5          & free-form & $63.9$ & $28.9$ & $80.0$ & $77.8$ & $68.9$ & $9.2$ \\
                 & concise structured & $76.1$ & $42.2$ & $95.6$ & $91.1$ & $75.6$ & $21.4$ \\
                 & structured & $81.1$ & $57.8$ & $91.1$ & $91.1$ & $84.4$ & $13.0$ \\
Gemini 3.1 Pro   & free-form & $11.1$ & $0.0$ & $4.4$ & $11.1$ & $28.9$ & $3.5$ \\
                 & concise structured & $48.9$ & $15.6$ & $64.4$ & $80.0$ & $35.6$ & $21.4$ \\
                 & structured & $51.1$ & $26.7$ & $57.8$ & $62.2$ & $57.8$ & $22.7$ \\
Claude Code      & free-form & $34.4$ & $15.6$ & $42.2$ & $24.4$ & $55.6$ & $7.4$ \\
                 & concise structured & $66.7$ & $40.0$ & $91.1$ & $73.3$ & $62.2$ & $12.0$ \\
                 & structured & $73.3$ & $60.0$ & $86.7$ & $71.1$ & $75.6$ & $6.8$ \\
\bottomrule
\end{tabular}
\end{table}

\section{Module attribution} \label{app:attribution}

A natural question about a multi-module reviewer such as \skepthical{} is which of its modules its detections come from. The rates of Table~\ref{tab:headline} credit \skepthical{} as a whole, so we attributed its astrophysics detections to modules by examining the reviews themselves. Each \skepthical{} review contains a general review and separate citation, mathematics, and numerics audit sections written by the specialist modules. For every injection of those three types, we recorded whether the corresponding specialist section identifies the injected error and whether the review identifies it anywhere, which is what the panel judges and Table~\ref{tab:headline} reports. This locates where the supporting evidence appears, although it does not say what would happen if a module were removed. The attribution was carried out by Claude in a single pass.

The attribution results are as follows. The citation audit identifies 11 of the 21 citation errors the review finds. The main failure mode is that the citing claim was never selected for checking. The numerics audit identifies 27 of the 36 numerics errors the review finds ($75\,\%$); its misses are statements never selected for a check, so claim selection is again its main failure. The mathematics audit identifies 7 of the 31 mathematics errors the review finds. Most mathematics detections thus come from the general-issues ensemble reading the whole paper. These results also argue against the explanation that \skepthical{} performs well merely because the benchmark's error types mirror its module names: one specialist section supplies most of its corresponding detections, one about half, and one a minority.

\section{Citation errors across the two corpora} \label{app:citations-transfer}

On astrophysics, \skepthical{} detects $46.7\,\%$ of the injected citation errors and no other system more than $26.7\,\%$. On ML, by contrast, two of the three structured single-LLM reviewers reach $57.8\,\%$ and \skepthical{} $53.3\,\%$. In this appendix, we examine the origin of this discrepancy. We recall that the single LLMs receive one call with the paper attached and no tools, so they entirely rely on their pre-trained knowledge to detect the citation injections. \skepthical{}, on the other hand, retrieves and reads the cited papers.

This difference may arise because the ML corpus contains more citation errors that can be caught from familiarity with the field alone, because the models are more familiar with the ML literature, or through a combination of the two. To separate these contributions, we classified each of the 45 injected citation errors of each corpus into three groups. The first group contains errors that a reader familiar with the field could reject without opening the cited paper, for instance a well-known method attributed to the wrong paper. The second contains errors for which the cited paper is needed, for instance a dropped condition or a changed number. The third contains errors that could fall either way. The classification was made in a single pass by Claude. The two corpora differ markedly in this respect: errors of the first group are more than three times as common on ML as on astrophysics (14 against 4 of the 45).

This difference in composition, however, explains only part of the single LLMs' gain. Taken together, the three structured single LLMs detect $16.7\,\%$ of the citation errors on astrophysics and $47.4\,\%$ on ML, and had the ML errors been distributed across the groups in the astrophysics proportions they would still have detected $39.6\,\%$. About a quarter of the gain therefore comes from the composition and the rest from higher detection within every group. The largest increase is in the first group, where knowing the field is enough. \skepthical{}'s rate, by contrast, is within a few points of its astrophysics value once the composition is accounted for. In short, the single LLMs gained on ML both because more of the errors can be caught from familiarity with the field and because they are more familiar with it, whereas \skepthical{}, which retrieves the cited paper in every case, changed little.

\section{Error coupling within a copy} \label{app:severity-coupling}

A reviewer reading an injected copy encounters four injected errors at once. A natural question to ask, which we examine in this appendix, is whether the detection of one error could depend on the severity of the other errors present in the same copy. A blatant error, in particular, may draw attention away from a subtle one. The most direct test would place each injected error in a copy of its own, which requires 12 injected copies per paper rather than three. For five papers alone this would amount to 60 additional reviewed copies, at a cost comparable to that of the entire astrophysics benchmark, so this experiment was not carried out.

Instead, we used the randomisation already present in the benchmark. Because severity was assigned independently for each error type, an error of a given severity appears beside three neighbours whose severities vary at random from copy to copy, so the effect of the neighbours can be read from the existing verdicts. For every verdict on every injected error of both corpora, we recorded whether the error was detected and the highest severity among the other three errors in its copy, taking together the verdicts of the systems of Table~\ref{tab:headline} and of the concise structured configurations of Appendix~\ref{app:concise}.

We find no measurable dependence on the neighbours. Over both corpora and all severities combined, the detection rate is $47$, $50$, and $45\,\%$ when the most severe neighbour is subtle, clear, and blatant. The difference between a blatant and a clear neighbour, the only comparison with enough errors behind it, is $-4.9\,\mathrm{pp}$ with a $95\,\%$ paper-level bootstrap interval of $[-11.8, +1.2]$, and the interval includes zero within each severity of the target error and within each corpus as well. A logistic model that controls for the target error's own severity, its type, the reviewer system, and the paper likewise finds no significant effect in either corpus.

We note that this test cannot determine whether the mere presence of other errors has an effect, because every copy carries four. Answering this question would require analysing the single-error copies described above, which is beyond the resources of this work.

\section{Inter-reviewer agreement} \label{app:inter-reviewer}

\begin{figure}[t]
\centering \includegraphics[width=0.99\linewidth]{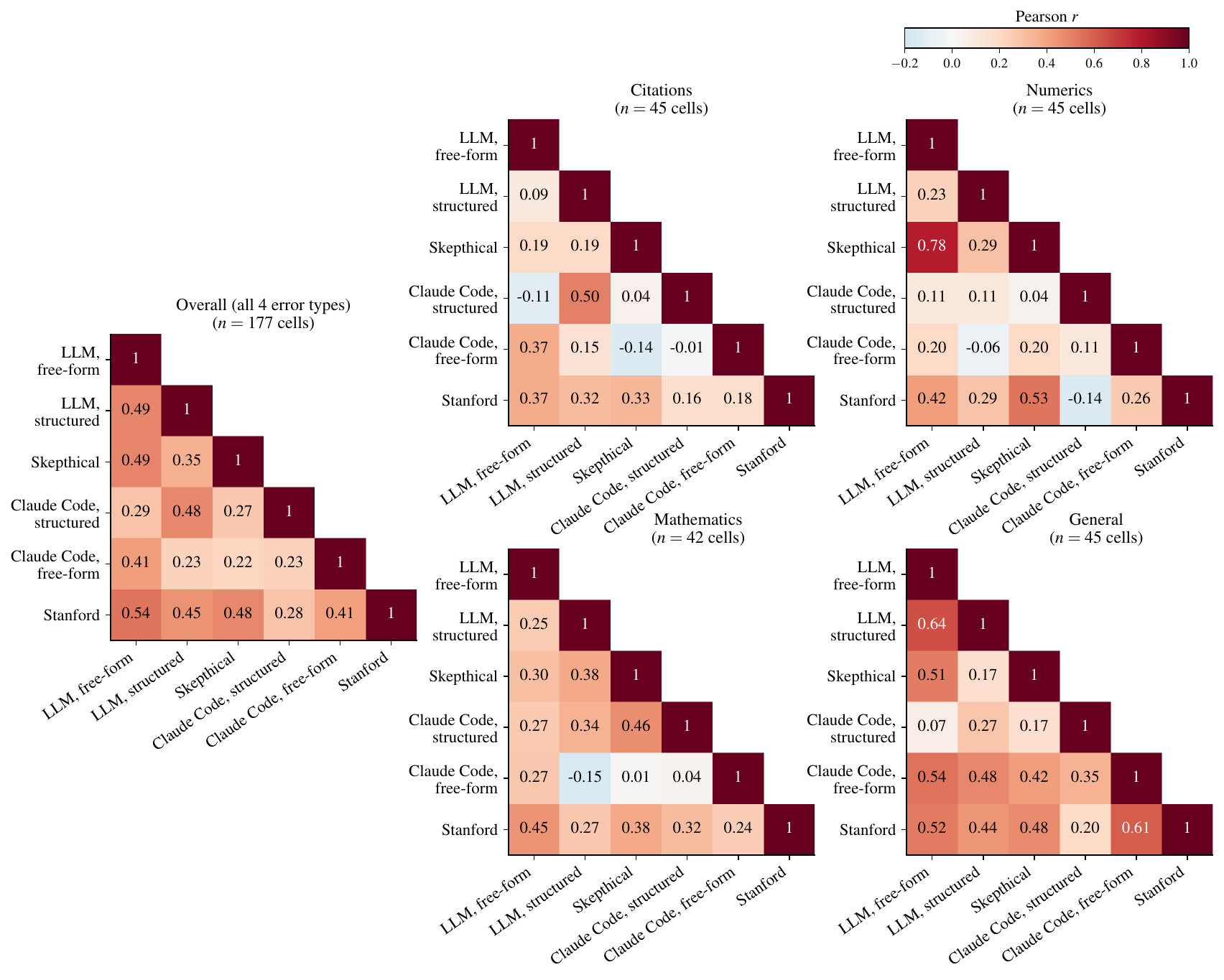} \caption{\textbf{Inter-reviewer Pearson $r$ matrices.} Left: overall $r$ across the six systems named in the text, on the astrophysics corpus. Right: per-error-type breakdown (citations and numerics on top, mathematics and general on the bottom). Diagonals are 1; off-diagonals are Pearson $r = \phi$, the Matthews correlation coefficient, for binary detection vectors. \skepthical{} $\leftrightarrow$ Claude Code (structured) $r = 0.27$ overall, $\approx 0.04$ on citations and numerics: two top-performing systems that detect different errors. Single-LLM (free-form) $\leftrightarrow$ Stanford $r = 0.54$ is the highest pair: the hosted Stanford reviewer behaves structurally like a free-form referee. The mathematics matrix is computed on $n = 42$ injections rather than $45$ because one paper in the corpus contains too few displayed equations to admit mathematics injections and is therefore absent from the mathematics row.} \label{fig:corr}
\end{figure}

Per-system detection rates (Table~\ref{tab:headline}) leave open the question of overlap: two systems with the same rate can detect the same errors or entirely different ones, and two systems with rates near zero can agree by chance. This appendix reports the pairwise agreement between reviewers on the astrophysics corpus, which shows which features of architecture and prompt lead different systems to detect the same errors. Figure~\ref{fig:corr} shows the pairwise Pearson $r$ between the binary per-injection detection vectors of six reviewer systems, overall and per error type: \skepthical{}, single-LLM (free-form), single-LLM (structured), Claude Code (free-form), Claude Code (structured), and the Stanford Agentic Reviewer. For the two single-LLM entries, an error counts as detected when any of the three providers detects it. The values lie between $0.2$ and $0.54$: the reviewers are heterogeneous, with similar overall detection rates but visibly different sets of detected errors.

The highest pair is single-LLM (free-form) and Stanford ($r = 0.54$), consistent with the hosted Stanford reviewer behaving like a free-form referee. Single-LLM (structured) and Claude Code (structured) are at $r = 0.48$: a shared prompt produces moderately similar detections even between a single call and an agentic loop with tools. \skepthical{} and Claude Code (structured) are at $r = 0.27$ overall and at $r \approx 0.04$ on citations and numerics, the error types on which each is the best of all systems: two top systems whose detections are essentially independent there.

This independence is above chance. Pearson $r$ on binary vectors measures agreement beyond what the two rates alone imply: two systems that detect $80$ and $91\,\%$ of the numerics errors already coincide on about $75\,\%$ of them for that reason, so $r \approx 0$ means that their detections coincide no more often than that, not that the systems disagree. As a check we also computed Cohen's $\kappa$. On the overall matrix it lies at or slightly below $r$ for every pair, by at most $0.10$, as expected for binary votes with unequal rates, and the conclusions drawn above with $r$ hold with $\kappa$ as well. \skepthical{} and structured Claude Code still agree only weakly overall ($\kappa = 0.27$) and hardly at all on citations and numerics ($\kappa \approx 0.04$). The single-LLM free-form and Stanford reviewers, the most similar pair under $r$, remain among the most similar under $\kappa$ ($0.48$, against a maximum of $0.49$).

Beyond per-injection agreement (Figure~\ref{fig:corr}), we also test convergence at the paper level. For each (paper, reviewer) we compute the reviewer's detection rate on that paper (out of $9$ or $12$ injections) and form the $15 \times 12$ rate matrix. The mean pairwise Pearson $r$ between reviewer columns is $+0.39$ ($n = 45$ pairs, restricted to the ten systems with non-negligible detection rates; the two open-weights baselines are excluded). Reviewers thus agree more on which papers are difficult in aggregate than on which specific injections each catches. The within-paper picture is the complement: a Pearson chi-square dispersion test on per-paper detection counts gives a mean ratio of $1.04$ across the same ten systems, indicating that within a paper detections behave approximately as independent Bernoulli draws across injections, so finding one injection does not strongly predict finding another one in the same paper.

\section{Compute and cost} \label{app:compute}

\subsection{Compute resources} \label{app:compute-resources}

The benchmark uses three classes of compute. (i) Hosted-API LLM calls to OpenAI, Anthropic, and Google serve every \skepthical{} agent (Table~\ref{tab:models}), the six single-LLM reviewers, the two Claude Code configurations, and the detection-matching and false-positive adjudication panels; the local CPU footprint of these calls is negligible (orchestration drivers pinned to 10 cores via \texttt{taskset}) since the work is dominated by external API latency. (ii) \skepthical{} itself runs as a Google Cloud Run worker (1~vCPU and 2~GiB of memory per task), with a per-paper wall time of approximately 24~minutes (Appendix~\ref{app:skepthical}). (iii) The two self-hosted open-weights baselines, OpenReviewer-8B (bfloat16, $\sim$16~GB) and the PaperAudit-Bench Qwen3-14B detector (bfloat16, $\sim$28~GB weights plus key-value cache), run sequentially on a single NVIDIA RTX PRO 6000 Blackwell Workstation Edition GPU (96~GB VRAM). The host workstation is an AMD Ryzen Threadripper PRO 9995WX (96 cores, 192 threads, 502~GiB RAM) with two such GPUs attached; only one of the two was used for the open-weights baselines.

\subsection{Cost estimates} \label{app:compute-cost}

The corpus run uses three billing modes: hosted-API metered calls (for the six single-LLM reviewers, every \skepthical{} agent, and the adjudication panels), Claude Code Max flat-rate subscription (for the two Claude Code configurations), and self-hosted or free-hosted operation with no API charge (the Stanford Agentic Reviewer, OpenReviewer-8B, and the PaperAudit-Bench detector).

\begin{table}[h]
\centering \footnotesize
\caption{Reviewer-side cost and median per-paper wall-clock by system on the 60 paper versions of the astrophysics corpus. ``USD per paper'' is corpus cost divided by the $60$ paper-version reviews. The single-LLM rows are averages over the three providers (Anthropic, OpenAI, Google) at the listed prompt level; their wall-clock is the median of the per-review times recorded at run time. The two Claude Code rows are subscription-equivalent at API rates with wall-clock from the per-turn token audit. Free-of-API-charge systems list \$$0$ reviewer-side cost; the open-weights wall-clocks are medians of the logged \texttt{wall\_seconds} on a single NVIDIA RTX PRO 6000 Blackwell GPU.}
\label{tab:cost}
\begin{tabular}{lrrr}
\toprule
System & Corpus cost (USD) & USD / paper & Wall-clock / paper \\
\midrule
\skepthical{}                                                  & $145$ & $2.41$ & $\sim\!24$ min \\
Single-LLM structured                                          & $33$  & $0.55$ & $\sim\!2$ min \\
Single-LLM free-form                                           & $27$  & $0.45$ & $\sim\!1$ min \\
Claude Code structured (sub.-equivalent)                       & $104$       & $1.73$ & $8.7$ min \\
Claude Code free-form (sub.-equivalent)                        & $20$        & $0.33$ & $1.4$ min \\
Stanford Agentic Reviewer (hosted, free)                       & $0$         & $0$    & $5$--$10$ min \\
OpenReviewer-8B (self-hosted)                                  & $0$         & $0$    & $\sim\!11$ s \\
PaperAudit-Bench detector (self-hosted)                        & $0$         & $0$    & $\sim\!36$ s \\
\bottomrule
\end{tabular}
\end{table}

Table~\ref{tab:cost}  shows the API spend incurred by the reviewer systems on the astrophysics corpus. Note that the Claude Code costs are estimates derived from the associated token usage (as they are run on a flat-rate subscription). Two observations follow. First, the per-paper cost spread across hosted-API systems is roughly sevenfold (\$$0.34$ to \$$2.41$), with \skepthical{} at the top end (\$$2.41$) reflecting its multi-agent design. Second, within Claude Code (model, sandbox, and tools held fixed), swapping the bare prompt for the structured operating instructions raises per-paper cost by $\sim\!5\times$ and per-paper wall-clock by $\sim\!6\times$.

These per-reviewer figures understate what the benchmark cost to build and run, since they exclude the injection of the errors and the adjudication of every review. Table~\ref{tab:cost-e2e} gives the complete logged API spend on each corpus, split into error injection, reviewer execution, detection adjudication, and false-positive adjudication, for the systems of Table~\ref{tab:headline} together with the concise structured configurations of Appendix~\ref{app:concise}.

\begin{table}[h]
\centering \footnotesize
\caption{Complete logged API cost (USD) by pipeline stage and corpus, for the systems of Table~\ref{tab:headline} together with the concise structured configurations of Appendix~\ref{app:concise}.}
\label{tab:cost-e2e}
\begin{tabular}{lrr}
\toprule
Stage & Astrophysics & ML \\
\midrule
Error injection              & $0.87$   & $40.04$  \\
Reviewer execution           & $351.59$ & $272.14$ \\
Detection adjudication       & $124.51$ & $73.82$  \\
False-positive adjudication  & $42.92$  & $23.27$  \\
Other (eligibility scan, re-reading) & $1.77$ & $1.89$ \\
\midrule
Total                        & $521.65$ & $411.17$ \\
Per paper                    & $34.78$  & $27.41$  \\
\bottomrule
\end{tabular}
\end{table}

\subsection{Assets and licences} \label{app:compute-assets}

The mechanical-verification stack uses Wolfram Engine (proprietary, with a free developer licence from Wolfram Research), SymPy (BSD-3-Clause), and \texttt{pint} (BSD). The two open-weights baselines are used under their published licences: OpenReviewer-8B (\texttt{maxidl/Llama-OpenReviewer-8B}) under the Llama 3.1 Community License, as a Meta-Llama-3.1-8B-Instruct fine-tune; and the PaperAudit-Bench Qwen3-14B detector (\texttt{mayiwen/PaperAudit\_Qwen3\_14B\_sft\_rl}) under Apache-2.0. The hosted LLMs from OpenAI, Anthropic, and Google (identifiers in Table~\ref{tab:models} and \S\ref{sec:systems}) are accessed under each provider's terms of service. The two 15-paper corpora consist of arXiv preprints used for research purposes only, with each paper credited individually in Appendix~\ref{app:corpus}; the papers themselves are not redistributed, and the release identifies them by their arXiv identifiers.

\end{document}